\pdfoutput=1

\documentclass[11pt]{article}

\usepackage[final]{acl}

\usepackage{times}
\usepackage{latexsym}

\usepackage[T1]{fontenc}

\usepackage[utf8]{inputenc}

\usepackage{microtype}

\usepackage{inconsolata}

\usepackage{graphicx}
\usepackage{booktabs}
\usepackage{multicol}
\usepackage{multirow}
\usepackage{makecell}

\title{TALEC: Teach Your LLM to Evaluate in Specific Domain with In-house Criteria by Criteria Division and Zero-shot Plus Few-shot}

\author{%
  *Kaiqi Zhang$^{1,2}$ \quad *Shuai Yuan$^{1}$ \quad *Honghan Zhao$^{1}$\\
  $^1$ByteDance China \quad $^2$University of Electronic Science and Technology of China\\
  \texttt{zhangkaiqi.zlk@gmail.com} \\
  \texttt{\{yuanshuai.irbaozi, zhaohonghan\}@bytedance.com} \\
}

\begin{document}
\maketitle
\footnotetext[0]{First submitted on June 25, 2024.}
\begin{abstract}
With the rapid development of large language models (LLM), the evaluation of LLM becomes increasingly important. Measuring text generation tasks such as summarization and article creation is very difficult. Especially in specific application domains (e.g., to-business or to-customer service), in-house evaluation criteria have to meet not only general standards (correctness, helpfulness and creativity, etc.) but also specific needs of customers and business security requirements at the same time, making the evaluation more difficult. So far, the evaluation of LLM in business scenarios has mainly relied on manual, which is expensive and time-consuming. In this paper, we propose a model-based evaluation method: \textbf{TALEC}, which allows users to flexibly set their own evaluation criteria, and uses in-context learning (ICL) to teach judge model these in-house criteria. In addition, we try combining zero-shot and few-shot to make the judge model focus on more information. We also propose a prompt paradigm and an engineering approach to adjust and iterate the shots ,helping judge model to better understand the complex criteria. We then compare fine-tuning with ICL, finding that fine-tuning can be replaced by ICL. TALEC demonstrates a strong capability to accurately reflect human preferences and achieves a correlation of over 80\% with human judgments, outperforming even the inter-human correlation in some tasks. The code is released in \href{https://github.com/zlkqz/auto_eval}{https://github.com/zlkqz/auto\_eval}.
\end{abstract}

\section{Introduction}

Automatically evaluating an outputted span of text from a model is difficult because of its uncertainty in text format and diversity of tasks. It is different from the other simple tasks like classification, which can simply evaluate outputs of models by splitting datasets. Automatic evaluation of a span of text usually uses a model-based (e.g., \citet{zheng2024judging}; \citet{jiang2023tigerscore}; \citet{wang2023pandalm}) or statistics-based (e.g., \citet{fu2023gptscore}; \citet{papineni2002bleu}); \citet{lin2004rouge}) method to evaluate. And it considers various standards (correctness, helpfulness and creativity, etc.) of the text. 

Since the birth of ChatGPT (\citet{ouyang2022training}) at the end of 2022, NLP research and development has officially entered the era of LLM. Although the R\&D of LLM is rapid, there is still a lack of available automatic evaluation methods for LLM. Especially in specific application domains (e.g., to-business or to-customer service), in-house evaluation criteria have to meet not only general standards (correctness, helpfulness and creativity, etc.) but also specific needs of customers and business security requirements at the same time, making the evaluation more difficult.

In this paper, we propose a model-based evaluation method: \textbf{TALEC}. TALEC focuses on evaluation in specific application scenarios. Besides, all the experiments and benchmark in this paper are related to our real application and are in the automobile field. TALEC allows users to flexibly set their own evaluation criteria, and uses in-context learning (ICL, \citet{brown2020language}) to teach judge model these in-house criteria. Our criteria can be viewed in Table \ref{tab:tab1}. We also propose an engineering approach to adjust and iterate the shots, which is splitting the dataset to "train", "eval" and "test" dataset. The "train" dataset is to find typical cases. Then we will provide these typical cases for the context as shots. The remaining two datasets is to help to adjust and iterate the shots and Verify the final result.

In addition, we find some problems when using shots which is written manually. Moreover, too many shots will also cause forgetting some former information and may exceeding context length limit. To solve this, We come up with a prompt paradigm and try combining zero-shot and few-shot to make the judge model focus on more information.

As all know, fine-tuning is a good way to make model adapt to a downstream task (\citet{devlin2018bert}), including evaluating the outputs of other models. But now, almost all SOTA LLMs (e.g., GPT-4 (\citet{achiam2023gpt})) are closed-source. Some people alternatively use weaker model like Llama (\citet{touvron2023llama}) to fine-tune a judge model. However, this method has its upper limit since the weakness of base model. Therefore, some people use SOTA models with ICL to judge. So we compare fine-tuning with ICL, finding that fine-tuning can be replaced by ICL.

In the end of this paper, we compare TALEC with the other automatic evaluation method and humans. TALEC demonstrates a strong capability to accurately reflect human preferences and
achieves a correlation of over 80\% with human judgments, outperforming many other methods and even the inter-human correlation in some tasks 

\begin{figure*}[htbp]
    \centering
    \includegraphics[width=0.85\linewidth]{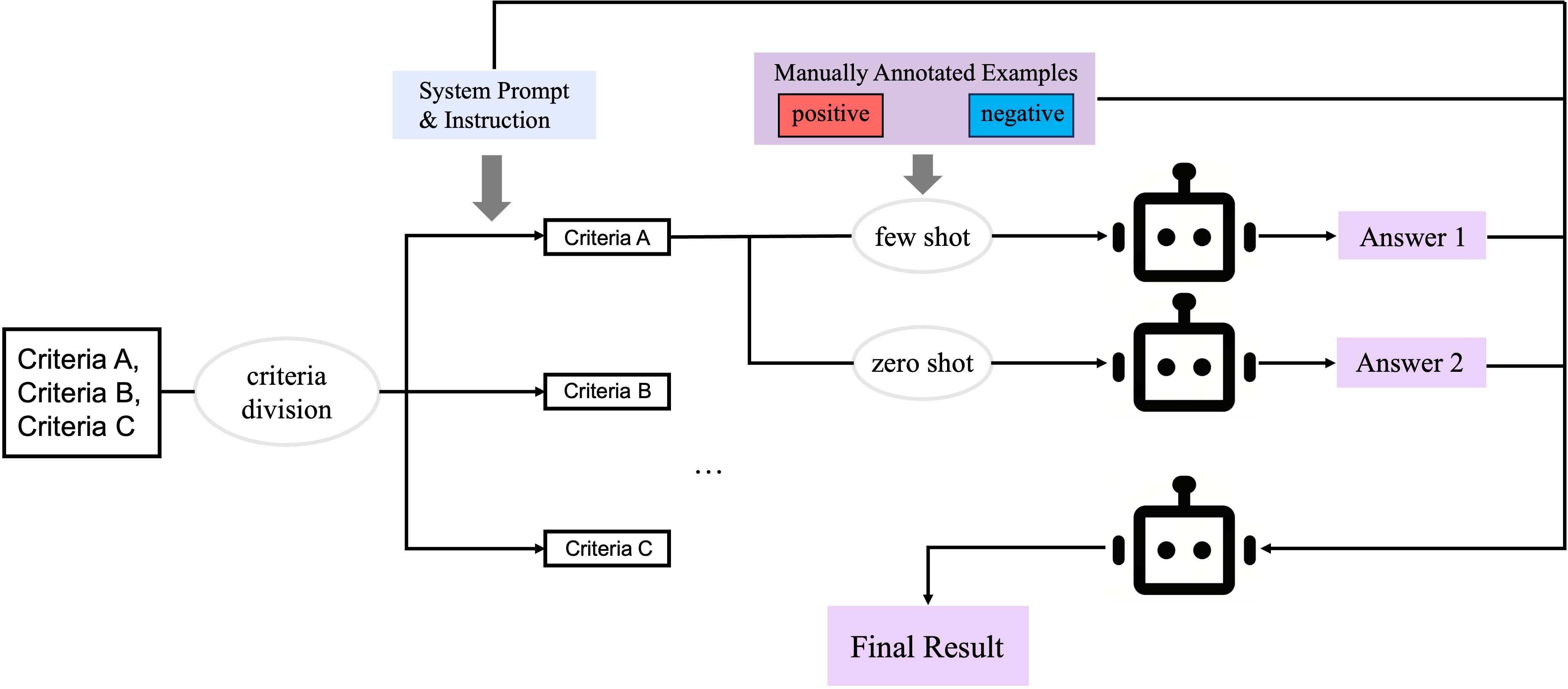}
    \caption{The overall process of TALEC}
    \label{fig:figure-1}
\end{figure*}

\section{Related Works}

The way of evaluating deep learning model has changed dramatically since the birth of ChatGPT. Before the this, various methods to automatically evaluate have been proposed. BLEU (\citet{papineni2002bleu}) and ROUGE (\citet{lin2004rouge}) calculate the similarity between output text and reference text. But restricted to LLMs' uncertainty in output text format and diversity of tasks, it is difficult to offer a good and proper reference text. GPTScore (\citet{fu2023gptscore}) uses perplexity (PPL, \citet{jelinek1977perplexity}) to evaluate output text based on former context, but recent research shows that there is no correlation between PPL and LLMs' long-text understanding ability (\citet{hu2024can}). MMLU (\citet{hendrycks2020measuring}), GPQA (\citet{rein2023gpqa}), C-Eval (\citet{huang2024c}) and some benchmark in SueprCLUE (\citet{xu2023superclue}) use multiple choice questions to evaluate. This method is simple and efficient, but it only focuses on model's knowledge and reasoning abilities, lacking of other abilities like instruction-following.

After the birth of ChatGPT, automatic evaluation methods become more explainable and pay more attention to the abilities of multiple dimensions of the model. MT-Bench (\citet{zheng2024judging}) uses GPT-4 to compare responses from tow different models and pay special attention to multi-turn dialogue ability of model. Besides, There are a lot of methods use fine-tuned model or aligned model to evaluate (\citet{jiang2023tigerscore}; \citet{wang2023pandalm}, etc.). Some methods focus on special abilities of LLM, such as LLM-EVAL (\citet{lin2023llm}) is a unified multidimensional automatic evaluation method for open-domain conversations with LLMs. TruthfulQA (\citet{lin2021truthfulqa}) focuses on hallucination of LLM. IFEval (\citet{zhou2023instruction}) designs some tasks to measure the instruction-following ability of LLM. HumanEval (\citet{chen2021evaluating}) and MBPP (\citet{austin2021program}) is benchmark to measure coding ability of LLM and they use pass@k as the final score. MATH (\citet{hendrycks2021measuring}) and GSM8K (\citet{cobbe2021training}) pay more attention to mathematical ability of LLM. And some methods use unique approach to evaluate, like BotChat (\citet{duan2023botchat}), which uses a approach similar to the Turing test to evaluate.

But all of them only use some general criteria (correctness, helpfulness and creativity, etc.) and have low correlation with human, making it unavailable in specific application domains. So we now  formally introduce our method: \textbf{TALEC}.

\section{Our Method: TALEC}

\subsection{Customized Business Evaluation Criteria}

TALEC is grounded in customizable, challenging, and adaptable evaluation criteria, distinguishing itself from conventional automatic methods. It allows users to flexibly set their own evaluation criteria, which maybe more difficult than some general criteria because the criteria may have to meet not only general standards but also specific needs of customers and business security requirements at the same time. In addition, it is hard to set the exact scale of the criteria, making it more difficult sometimes. Even for a human evaluator, a series of practices and quality inspections is needed. 

In this paper, we do experiments on four distinct tasks and ten customized labels. The tasks include: 

\textbf{Sentiment Analysis.} \ Given a comment, determine the type of sentiment based on the textual information and provide the reason for the judgment.

\textbf{Knowledge QA.} \ Given a knowledge question in the automobile field, provide a detailed answer to this question.

\textbf{Search QA.} \ Given a piece of reference information from search engines and a related question, answer the question based on the reference information. 

\textbf{Title Generation.} \ Given an article and some complex requirements, generate main title and subtitle that is based on the article and completely meets the requirements.

Labels and their descriptions can be viewed in Table \ref{tab:tab1}. The labels can be divided into two categories: acceptable labels(score=1) and unacceptable labels(score=0). If any unacceptable label appears, the final sore will be 0. If unacceptable label does not appear but any acceptable label appears, the final sore will be 1. If no labels appears, the final sore will be 2 (full score). 

\begin{table*}[]
\resizebox{\textwidth}{!}{%
\begin{tabular}{@{}l|l|c@{}}
\toprule
\textbf{Label Name}             & \textbf{Description}                                                                                         & \multicolumn{1}{l}{\textbf{Score}} \\ \midrule
Not Meeting the Requirements                  & Answer failed to strictly enforce the requirements.                             & 0                                  \\
Incorrect Answer/Unrelated Matching Results    & Containing incorrect information or mismatched responses.                                & 0                                  \\
Refusal to Answer               & The model explicitly shows a refusal to answer when it can.       & 0                                  \\
Untranslated Text               & Large portions of non-Chinese language in responses.         & 0                                  \\
Confusing Answers                & Answers contain messy code or content that interferes with reading.
       & 0                                  \\
External Links or Diversions & External links or obvious diversionary behavior in the answer.                                    & 0                                  \\
Stiffness              & The response lacked anthropomorphic expression.                                    & 1                                  \\
Repetitive Expression          & Repeated expressions in the answer.                                               & 1                                  \\
Subject Imprecision               & Redundant content or unclear subject in the response
 & 1                                  \\
Incomplete Answers               & Failure to address all needs in the directive                                                & 1                                  \\ \bottomrule
\end{tabular}%
}
\caption{Labels to be evaluated. The labels can be divided into two categories: acceptable labels(score=1) and unacceptable labels(score=0). If any unacceptable label appears, the final sore will be 0. If unacceptable label does not appear but any acceptable label appears, the final sore will be 1. If no labels appears, the final sore will be 2 (full score). Note that we don't use our model-based method to judge the word count because LLM can't accurately count the number of words.}
\label{tab:tab1}
\end{table*}



\subsection{Assumption of TALEC}

Many automatic evaluation only rely on the ability of the model itself, without any knowledge injection and teaching based on concrete problems and examples. Actually, even for a human evaluator, a series of practices with concrete examples and quality inspections is needed to learn our evaluation criteria. Therefore, the point is treating judge model like a human evaluator. What we need to do is teaching the judge model repeatedly and patiently with concrete examples. In order to do this, we simulate this practice-quality inspection cycle process by splitting the dataset to "train", "eval", "test" dataset and adding manual adjustment. The details of this simulation will be displayed in Section 3.3.

\subsection{TALEC}

We introduce TALEC, a novel automatic evaluation framework that leverages SOTA model like GPT-4 to evaluate an outputted span of text from a model. TALEC mainly uses ICL to teach judge model the customized evaluation criteria. The overall process of TALEC can be viewed in Figure \ref{fig:figure-1}. We will introduce several key points of TALEC below.

\textbf{Engineering Approach to Adjust and Iterate the Shots.} \ We split the dataset to "train", "eval", "test" dataset. The "train" dataset is to find typical cases. Then we will provide these typical cases for the context as shots. The "eval" dataset is to help manual optimization. The overall process can be listed as: \textbf{(1).} Find a some typical cases by feeling, then write the reasons why the cases are wrong, and regard them as the first version of shots. \textbf{(2).} Use this version of shots to run and gather statistics on "eval" dataset. \textbf{(3).} adjust the shots (add/delete/modify) based on the results on "eval" dataset. \textbf{(4).} Repeat the above process until you get a better results on "eval" dataset. After this process, use the final version of shots to run and gather statistics on "test" dataset, to verify the effectiveness of the shots.

\textbf{Criteria Division.} \ In our customized criteria, there are 10 labels per task. Each label has several positive and negative shots. So too many shots may cause exceeding context length limit. Criteria division will solve this problem. It is to divide the overall criteria to label granularity. For example, we assume there are 10 labels. Normally, we will let GPT-4 determine whether these 10 labels exist at one time. But criteria division is to let GPT-4 judge 10 times, and only evaluate one label each time. In addition, we also find that despite a 10-fold increase in cost, criteria division result in  greatly improvement in judging on almost all the labels. We will discuss the improvement in Section 4.1. 

\textbf{Prompt Paradigm.} \ ICL is a good way to teach judge model the in-house criteria. However, it will cause some problems. When injecting manually-written shots into the context of a model, the model will not only try to understand the shots but also imitate the writing style of shots. This imitation may make the model ignore some key information and drop its Chain of Thought (CoT, \citet{wang2023chain}) ability. The prompt paradigm can be listed as: \textbf{(1).} Repeat the description of a label before judge. \textbf{(2).} Try not to use transitive ( or) progressive words such as "and", "but", "however" in the first half of the judge reason. \textbf{(3).} Try to keep the positive and negative shots consistent in formatting, especially in the first half. We will verify the effectiveness of our prompt paradigm in section 5.

\textbf{Combine Zero-shot with Few-shot.} \ This approach is to compensate for the model's omission of key information. As Figure \ref{fig:figure-1} shows, the model will make two completely independent judgments, one is zero-shot judge and another one is few-shot judge. Then we will connect system Prompt, Shots, answer outputted by zero-shot, answer outputted by few-shot to get a new context and use this context to judge again. Then we will get the final result. The prompt template can be viewed in Figure \ref{fig:figure-4}. Ablation experiment results of this approach is shown in Section 4.2.

\section{Ablation Experiment}

We verify the effectiveness of the approaches mentioned above. Note that we use different variants of GPT-4 (gpt-4-0613, gpt-4-32k-0613 and gpt-4-0125-preview) to suit different contexts length. The baseline of these experiments is called \textbf{Standard Prompt Paradigm}, which uses engineering approach to adjust and iterate the shots and applies criteria division and our prompt paradigm. However, Standard Prompt Paradigm does not combine zero-shot with few-shot, it only uses a conventional ICL approach. 

During a series of experiments, we find that "Incorrect Answer/Unrelated Matching Results" label in Knowledge QA task is very special. A uniformly formatted few-shot would instead negatively affect the label, which distinguishes itself from the others. We guess it is because the errors in the answer of the Knowledge QA task may be evenly distributed throughout the answer. Shots is useless in this scenario because it is difficult to help localize the error information. Furthermore, uniformly formatted shots will further limit model's ability to find error information.

\subsection{Criteria Division}

Our evaluation methodology employs large language models, with a constraint on their maximum context length. Generally, this approach suffices for most tasks. However, in instances where lengthy prompts and responses are required, such as in the context of few-shot article generation, a single instance can extend to 1500-2000 tokens, thereby posing a limitation on the context length. Furthermore, when evaluations involve intricate tasks with multiple dimensions, the accuracy of the assessment may be negatively affected.

To address these challenges, we strategically decompose the customized evaluation criteria into distinct components, primarily by segmenting the evaluation dimensions. This approach enables the model to concentrate more on each criterion, thereby streamlining the evaluation process and enhancing the outcomes. By associating each problem label with its corresponding few-shot examples and inputting them into the model, we bypass the need for a single evaluation of all labels. Additionally, this method effectively alleviates the problem of insufficient context.

We have attempted to compare the following two experimental setups:

\textbf{Standard Prompt Paradigm(Division). } Individual evaluation dimensions are fragmented into distinct criteria, which are separately fed into the Judge model. The model's output is aggregated multiple times for each criterion before calculating the overall score.

\textbf{Non-division. } The complete set of evaluation criteria, along with their associated shots, is simultaneously fed into the model, enabling it to produce the final score directly without sequential processing.

Table \ref{tab:tab2} shows the results. In the tasks of Sentiment Analysis, Search QA, and Title Generation, the evaluation results of the criteria division method significantly outperform those of non-division. Conversely, in the task of knowledge QA, the situation is reversed. We found that the primary cause of this discrepancy lies in the "Incorrect Answer/Unrelated Matching Results" label as shown in Table \ref{tab:tab3}. As previously mentioned, this is because when criteria division is applied, the prompt format is much more distinct than when not divided, making the model more susceptible to format imitation, thereby overlooking practical information.

\begin{table}[htbp]
\centering
\scalebox{0.8}{%
\begin{tabular}{clcc}
\hline
\multicolumn{1}{l|}{\multirow{2}{*}{Method}} & \multicolumn{1}{l|}{\multirow{2}{*}{Task}} & \multicolumn{2}{c}{Spearman}                       \\ \cline{3-4} 
\multicolumn{1}{l|}{}                        & \multicolumn{1}{l|}{}                      & \multicolumn{1}{c|}{eval}   & test                 \\ \hline
\multicolumn{1}{c|}{\multirow{4}{*}{\shortstack{Standard Prompt \\ Paradigm \\ (Division)}}}     & \multicolumn{1}{l|}{Sentiment Analysis}      & \multicolumn{1}{c|}{0.9579} & 0.9693               \\
\multicolumn{1}{c|}{}                        & \multicolumn{1}{l|}{Knowledge QA}          & \multicolumn{1}{c|}{0.4945} & 0.5063               \\
\multicolumn{1}{c|}{}                        & \multicolumn{1}{l|}{Search QA}             & \multicolumn{1}{c|}{0.8263} & 0.8487                \\
\multicolumn{1}{c|}{}                        & \multicolumn{1}{l|}{Title Generation}      & \multicolumn{1}{c|}{0.9207} & 0.9006               \\ \hline
\multicolumn{1}{c|}{\multirow{4}{*}{\shortstack{Non-division}}}     & \multicolumn{1}{l|}{Sentiment Analysis}      & \multicolumn{1}{c|}{0.3735} & 0.2905               \\
\multicolumn{1}{c|}{}                        & \multicolumn{1}{l|}{Knowledge QA}          & \multicolumn{1}{c|}{0.5398} & 0.4251               \\
\multicolumn{1}{c|}{}                        & \multicolumn{1}{l|}{Search QA}             & \multicolumn{1}{c|}{0.689} & 0.4691              \\
\multicolumn{1}{c|}{}                        & \multicolumn{1}{l|}{Title Generation}      & \multicolumn{1}{c|}{0.3763} & 0.4296               \\ \hline

\end{tabular}%
}
\caption{Comparison between criteria division and non-division. The details can be found in Table \ref{tab:tab102} and Table \ref{tab:tab106}.}
\label{tab:tab2}
\end{table}

\begin{table}[htbp]
\centering
\scalebox{0.4}{%
\begin{tabular}{c|c|c|c|c|c|c}
\hline \multirow{2}{*}{ Method } & \multirow{2}{*}{ task } & \multirow{2}{*}{ label } &  \multicolumn{2}{|c|}{ Acc } & \multicolumn{2}{|c}{ F1/Precision/Recall } \\
\cline{4-7} & & & eval & test & eval & test \\
\hline  \shortstack{Standard Prompt \\ Paradigm \\ (Division)}  & 
\shortstack{Knowledge \\ QA} & \shortstack{Incorrect Answer \\ /Unrelated \\ Matching Results}  & 0.8298 & 0.7737 &  0.52/0.3611/0.9286  &  0.2439/0.1667/0.4545  \\
\hline  Non-division  & 
\shortstack{Knowledge \\ QA} & \shortstack{Incorrect Answer \\ /Unrelated \\ Matching Results}  & 0.9149 & 0.8321 &  0.5714/0.5714/0.5714  &  0.1481/0.125/0.1818  \\
\hline
\end{tabular}
}
\caption{The detailed score comparison on "Incorrect Answer/Unrelated Matching Results" label in Knowledge QA task. The experimental setups are the same as in Table \ref{tab:tab2}.}
\label{tab:tab3}
\end{table}

\subsection{Combine Zero-shot with Few-shot}

We said above that the imitation to text format in shots may make the judge model ignore some key information. So we try injecting zero-shot judge to avoid the impact of shots. The process can be viewed in Figure \ref{fig:figure-1} and the prompt can be viewed in Figure \ref{fig:figure-4}.

We compare two experimental setups to verify the effectiveness of this approach:

\textbf{Standard Prompt Paradigm(Single-turn wo Zero-shot).} \ Use the shots obtained from our engineering approach and inject the shots into context to judge. This approach only judges one case in a single-turn.

\textbf{Multi-turn with Zero-shot.} \ As shown in Figure \ref{fig:figure-1}, the model will make two completely independent judgments, one is zero-shot judge and another one is few-shot judge. Then we will connect system Prompt, Shots, answer outputted by zero-shot, answer outputted by few-shot to get a new context and use this context to judge again. Then we will get the final result.

Table \ref{tab:tab4} shows comparative results of the two approach.

\begin{table}[htbp]
\centering
\scalebox{0.8}{%
\begin{tabular}{clcc}
\hline
\multicolumn{1}{l|}{\multirow{2}{*}{Method}} & \multicolumn{1}{l|}{\multirow{2}{*}{Task}} & \multicolumn{2}{c}{Spearman}                       \\ \cline{3-4} 
\multicolumn{1}{l|}{}                        & \multicolumn{1}{l|}{}                      & \multicolumn{1}{c|}{eval}   & test                 \\ \hline
\multicolumn{1}{c|}{\multirow{4}{*}{\shortstack{Standard Prompt \\ Paradigm \\ (Single-turn \\ wo Zero-shot)}}}     & \multicolumn{1}{l|}{Sentiment Analysis}      & \multicolumn{1}{c|}{0.9579} & 0.9693               \\
\multicolumn{1}{c|}{}                        & \multicolumn{1}{l|}{Knowledge QA}          & \multicolumn{1}{c|}{0.4945} & 0.5063               \\
\multicolumn{1}{c|}{}                        & \multicolumn{1}{l|}{Search QA}             & \multicolumn{1}{c|}{0.8263} & 0.8487                \\
\multicolumn{1}{c|}{}                        & \multicolumn{1}{l|}{Title Generation}      & \multicolumn{1}{c|}{0.9207} & 0.9006               \\ \hline
\multicolumn{1}{c|}{\multirow{4}{*}{\shortstack{Multi-turn \\ with Zero-shot}}}     & \multicolumn{1}{l|}{Sentiment Analysis}      & \multicolumn{1}{c|}{0.9536} & 0.9485               \\
\multicolumn{1}{c|}{}                        & \multicolumn{1}{l|}{Knowledge QA}          & \multicolumn{1}{c|}{0.8597} & 0.7089               \\
\multicolumn{1}{c|}{}                        & \multicolumn{1}{l|}{Search QA}             & \multicolumn{1}{c|}{0.8574} & 0.9438              \\
\multicolumn{1}{c|}{}                        & \multicolumn{1}{l|}{Title Generation}      & \multicolumn{1}{c|}{0.915} & 0.9206               \\ \hline
\end{tabular}%
}
\caption{Comparison between the results of single-turn without zero-shot and multi-turn with zero-shot. The details can be found in Table \ref{tab:tab102} and Table \ref{tab:tab103}.}
\label{tab:tab4}
\end{table}

\begin{table}[htbp]
\centering
\scalebox{0.4}{%
\begin{tabular}{c|c|c|c|c|c|c}
\hline \multirow{2}{*}{ Method } & \multirow{2}{*}{ task } & \multirow{2}{*}{ label } &  \multicolumn{2}{|c|}{ Acc } & \multicolumn{2}{|c}{ F1/Precision/Recall } \\
\cline{4-7} & & & eval & test & eval & test \\
\hline  \shortstack{Standard Prompt \\ Paradigm \\ (Single-turn wo Zero-shot)}  & 
\shortstack{Knowledge \\ QA} & \shortstack{Incorrect Answer \\ /Unrelated \\ Matching Results}  & 0.8298 & 0.7737 &  0.52/0.3611/0.9286  &  0.2439/0.1667/0.4545  \\
\hline  Multi-turn with Zero-shot  & 
\shortstack{Knowledge \\ QA} & \shortstack{Incorrect Answer \\ /Unrelated \\ Matching Results}  & 0.9645 & 0.9051 &  0.8148/0.8462/0.7857  &  0.5185/0.4375/0.6364  \\
\hline
\end{tabular}
}
\caption{The detailed score comparison on Incorrect Answer/Unrelated Matching Results label in Knowledge QA task. The experimental setups are the same as in Table \ref{tab:tab4}.}
\label{tab:tab4.2}
\end{table}

As you can see in Table \ref{tab:tab4}, multi-turn with zero-shot approach can improve scores on all task. Especially in "Incorrect Answer/Unrelated Matching Results" label of Knowledge QA, Table \ref{tab:tab4.2} shows greatly improvement in this label. 

\subsection{SFT vs ICL}

Previous automated evaluation methods have opted to incorporate knowledge through Supervised Fine-Tuning. However, state-of-the-art models such as GPT-4, due to their proprietary nature, cannot be subjected to SFT to further enhance their performance. Therefore, methods utilizing SFT are compelled to resort to relatively weaker open-source models as a foundation, which may potentially impact the evaluation results. Consequently, we have chosen to introduce knowledge using In-Context Learning.

We validate the differences in knowledge introduction using SFT and ICL methods respectively, by fine-tuning Qwen-72B-Chat model (\citet{bai2023qwen}) and comparing the effects. We have established three experimental setups using three different models: 

\textbf{Standard Prompt Paradigm(GPT4 + ICL).} Using GPT-4 as the evaluation model with in-context learning.

\textbf{Qwen-72B-Chat + ICL.} Using Qwen-72B-Chat as the evaluation model with in-context learning.

\textbf{Qwen-72B-Chat + SFT.} Fine-tuning Qwen-72B-Chat as the evaluation model. We construct a dataset composed of 179 manually annotated high-quality data specific to the aforementioned four tasks, 100 open-source evaluation data obtained from the TigerScore dataset, and 300 open-source general data. We have trained Qwen-72B-Chat for one epoch on this dataset.

The results are shown in Table \ref{tab:tab5}. It can be observed that GPT-4 with in-context learning outperforms the method based on Qwen-72B-Chat across all tasks. Meanwhile, Qwen-72B-Chat integrated with in-context learning exhibits comparable performance to Qwen-72B-Chat combined with SFT on two tasks, surpasses the latter on one task, and underperforms on the other task. This suggests that in-context learning can achieve results similar to SFT on LLM, and to a certain extent, can serve as a substitute for SFT. Furthermore, the in-context learning approach can be applied to state-of-the-art proprietary LLM with superior performance, thereby yielding enhanced results.

\begin{table}[htbp]
\centering
\scalebox{0.7}{%
\begin{tabular}{clcc}
\hline
\multicolumn{1}{l|}{\multirow{2}{*}{Method}} & \multicolumn{1}{l|}{\multirow{2}{*}{Task}} & \multicolumn{2}{c}{Spearman}                       \\ \cline{3-4} 
\multicolumn{1}{l|}{}                        & \multicolumn{1}{l|}{}                      & \multicolumn{1}{c|}{eval}   & test                 \\ \hline
\multicolumn{1}{c|}{\multirow{4}{*}{\shortstack{Standard Prompt Paradigm \\ (GPT4 + ICL)}}}     & \multicolumn{1}{l|}{Sentiment Analysis}      & \multicolumn{1}{c|}{0.9579} & 0.9693               \\
\multicolumn{1}{c|}{}                        & \multicolumn{1}{l|}{Knowledge QA}          & \multicolumn{1}{c|}{0.4945} & 0.5063               \\
\multicolumn{1}{c|}{}                        & \multicolumn{1}{l|}{Search QA}             & \multicolumn{1}{c|}{0.8263} & 0.8487                \\
\multicolumn{1}{c|}{}                        & \multicolumn{1}{l|}{Title Generation}      & \multicolumn{1}{c|}{0.9207} & 0.9006               \\ \hline
\multicolumn{1}{c|}{\multirow{4}{*}{\shortstack{Qwen-72B-Chat + ICL}}}     & \multicolumn{1}{l|}{Sentiment Analysis}      & \multicolumn{1}{c|}{0.7676} & 0.7578               \\
\multicolumn{1}{c|}{}                        & \multicolumn{1}{l|}{Knowledge QA}          & \multicolumn{1}{c|}{0.2969} & 0.1766               \\
\multicolumn{1}{c|}{}                        & \multicolumn{1}{l|}{Search QA}             & \multicolumn{1}{c|}{0.519} & 0.4513              \\
\multicolumn{1}{c|}{}                        & \multicolumn{1}{l|}{Title Generation}      & \multicolumn{1}{c|}{0.1585} & 0.1392               \\ \hline
\multicolumn{1}{c|}{\multirow{4}{*}{\shortstack{Qwen-72B-Chat + SFT}}}     & \multicolumn{1}{l|}{Sentiment Analysis}      & \multicolumn{1}{c|}{0.4565} & 0.4211               \\
\multicolumn{1}{c|}{}                        & \multicolumn{1}{l|}{Knowledge QA}          & \multicolumn{1}{c|}{0.1886} & 0.2444               \\
\multicolumn{1}{c|}{}                        & \multicolumn{1}{l|}{Search QA}             & \multicolumn{1}{c|}{0.5705} & 0.4772               \\
\multicolumn{1}{c|}{}                        & \multicolumn{1}{l|}{Title Generation}      & \multicolumn{1}{c|}{0.2738} & 0.2671               \\ \hline
\end{tabular}%
}
\caption{Comparison between GPT4 with in-context learning, Qwen-72B-Chat with in-context learning and Qwen-72B-Chat SFT without in-context learning.The details can be found in Table \ref{tab:tab102}, Table \ref{tab:tab104} and Table \ref{tab:tab105}.}
\label{tab:tab5}
\end{table}

\section{Prompt Engineering}

\subsection{Repeat Descriptions of Evaluation Criteria}

In the system prompt, we explicitly offer descriptions of the evaluation criteria to help model better understand the criteria. However, we noticed that the model occasionally failed to recall the previously provided descriptions because of very long context caused by too many shots.

For instance, the requirements for some generation tasks include word count specifications. However, due to the token-based tokenizer structure of LLLMs, they can't accurately count the number of words. Consequently, we employed a rule-based approach to judge this aspect, ensuring more precise assessments without relying on the model's limitations. To clarify this, we clarified in the prompt that the model should disregard word count requirements. Surprisingly, the model continued to consider it in its evaluations.

\begin{figure*}[htbp]
    \centering
    \includegraphics[width=0.55\linewidth]{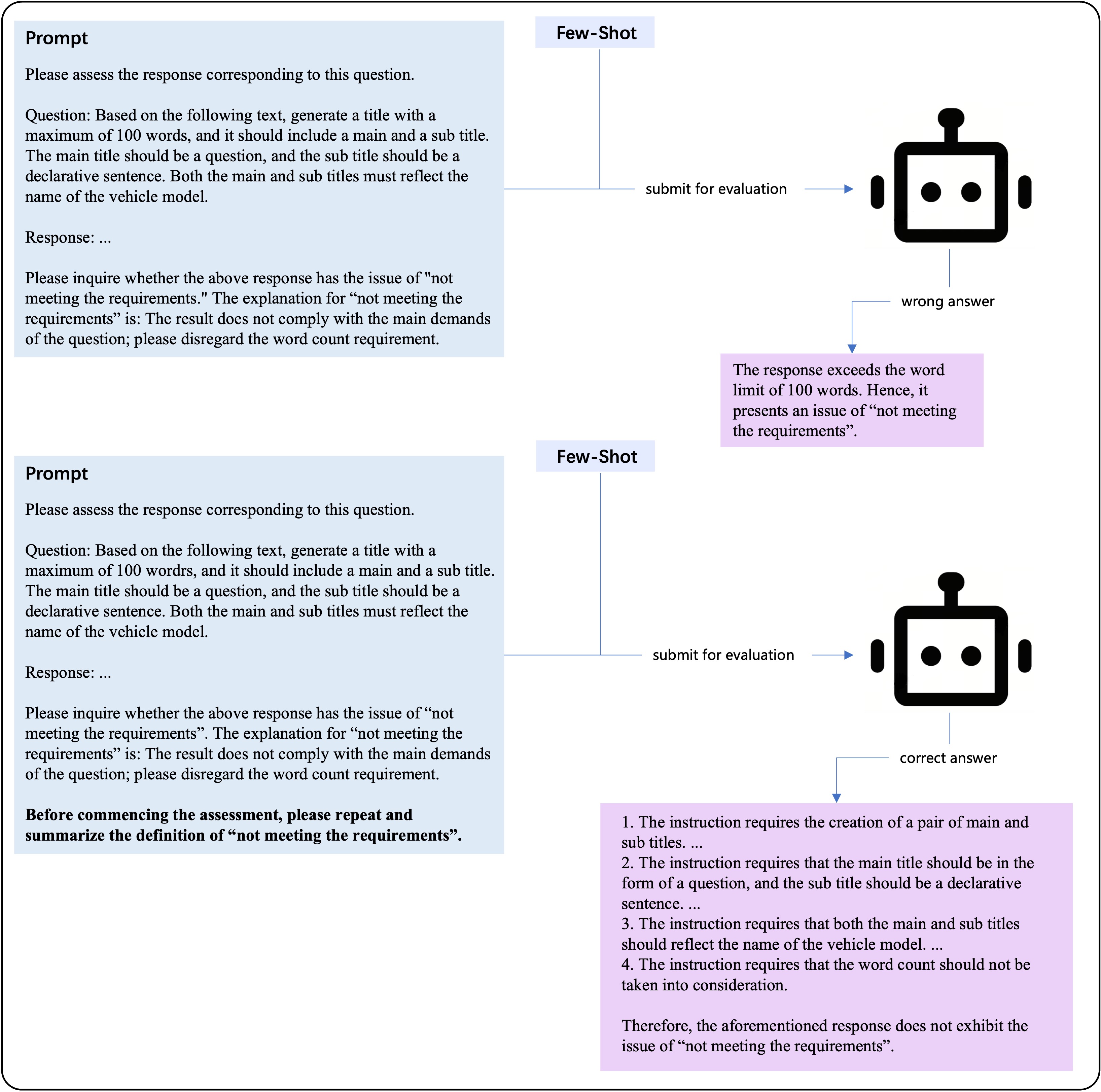}
    \caption{This figure illustrates an instance where the model fails to consider the descriptions within the System Prompt, juxtaposed with a contrasting example where the model redundantly repeats the descriptions before proceeding to evaluation.}
    \label{fig:figure-2}
\end{figure*}

To mitigate this issue, we introduce a novel approach where the model is prompted to recurrently summarize and reiterate the interpretation from the system prompt before delivering its evaluation, as depicted in Figure \ref{fig:figure-2}. 

To validate the efficacy of this method, we executed two experiments employing distinct strategies: 

\textbf{Standard Prompt Paradigm(Repeat descriptions).} We required the model to reiterate the descriptions prior to evaluation, maintaining a consistent format for the shots.

\textbf{Non-repetition.} We adopted a more informal format for the shot composition, eliminating the need for the model to reiterate the descriptions.

Table \ref{tab:tab6} shows the results. The methodology of repeating descriptions slightly outperforms the non-repetitive approach in three tasks, and significantly surpasses the latter in the remaining task. Experimental results indicate that repeating the descriptions of evaluation criteria prior to each label evaluation can assist LLM in better understanding the requirement of the evaluation, thereby enhancing accuracy.

\begin{table}[htbp]
\centering
\scalebox{0.7}{%
\begin{tabular}{clcc}
\hline
\multicolumn{1}{l|}{\multirow{2}{*}{Method}} & \multicolumn{1}{l|}{\multirow{2}{*}{Task}} & \multicolumn{2}{c}{Spearman}                       \\ \cline{3-4} 
\multicolumn{1}{l|}{}                        & \multicolumn{1}{l|}{}                      & \multicolumn{1}{c|}{eval}   & test                 \\ \hline
\multicolumn{1}{c|}{\multirow{4}{*}{\shortstack{Standard Prompt \\ Paradigm \\ (Repeat descriptions)}}}     & \multicolumn{1}{l|}{Sentiment Analysis}      & \multicolumn{1}{c|}{0.9579} & 0.9693               \\
\multicolumn{1}{c|}{}                        & \multicolumn{1}{l|}{Knowledge QA}          & \multicolumn{1}{c|}{0.4945} & 0.5063               \\
\multicolumn{1}{c|}{}                        & \multicolumn{1}{l|}{Search QA}             & \multicolumn{1}{c|}{0.8263} & 0.8487                \\
\multicolumn{1}{c|}{}                        & \multicolumn{1}{l|}{Title Generation}      & \multicolumn{1}{c|}{0.9207} & 0.9006               \\ \hline
\multicolumn{1}{c|}{\multirow{4}{*}{Non-repetition}}     & \multicolumn{1}{l|}{Sentiment Analysis}      & \multicolumn{1}{c|}{0.9553} & 0.9658               \\
\multicolumn{1}{c|}{}                        & \multicolumn{1}{l|}{Knowledge QA}          & \multicolumn{1}{c|}{0.4911} & 0.4823               \\
\multicolumn{1}{c|}{}                        & \multicolumn{1}{l|}{Search QA}             & \multicolumn{1}{c|}{0.8208} & 0.838               \\
\multicolumn{1}{c|}{}                        & \multicolumn{1}{l|}{Title Generation}      & \multicolumn{1}{c|}{0.6618} & 0.6815               \\ \hline
\end{tabular}%
}
\caption{Comparison of the experimental results for repeating descriptions and non-repetition. The details can be found in Table \ref{tab:tab102} and Table \ref{tab:tab107}.}
\label{tab:tab6}
\end{table}

\subsection{Standardize the Format of Examples}

It is universally recognized that large models possess Chain of Thought (CoT) capabilities. The prudent use of CoT enables the model to provide a detailed explanation before delivering the final answer, thereby substantially improving the accuracy of responses. However, in our approach, the incorporation of 'shots' has instigated a problem. As models mimic the structure of the shots in their output, an informal arrangement of shots could potentially cause the model to prematurely conclude the answer without a comprehensive explanation.



In the previously mentioned example, it seems that the model's Chain of Thought(CoT) capability is activated by initially offering an explanation. However, the model actually determines the output label at the outset due to the varied formats employed in the construction of positive and negative examples, particularly the inclusion of adversative phrases preceding the negative examples. The explanation is subsequently appended following the decision on the label, leading to an inversion of cause and effect.

Hence, we propose that in the realm of prompt engineering, the utilization of a consistent format for both positive and negative examples is crucial. This should be accompanied by a reduction in the use of adversative expressions. The objective is to postpone the revelation of the answer, thereby enabling the model to provide a comprehensive explanation prior to presenting the ultimate response at the conclusion.

Table \ref{tab:tab7} shows the results. It can be observed that the method of employing a standardized format for both positive and negative instances surpasses the method of using arbitrary formats in evaluation results across all tasks.

\begin{table}[htbp]
\centering
\scalebox{0.8}{%
\begin{tabular}{clcc}
\hline
\multicolumn{1}{l|}{\multirow{2}{*}{Method}} & \multicolumn{1}{l|}{\multirow{2}{*}{Task}} & \multicolumn{2}{c}{Spearman}                       \\ \cline{3-4} 
\multicolumn{1}{l|}{}                        & \multicolumn{1}{l|}{}                      & \multicolumn{1}{c|}{eval}   & test                 \\ \hline
\multicolumn{1}{c|}{\multirow{4}{*}{Arbitrariness}}     & \multicolumn{1}{l|}{Sentiment Analysis}      & \multicolumn{1}{c|}{0.9551} & 0.9515               \\
\multicolumn{1}{c|}{}                        & \multicolumn{1}{l|}{Knowledge QA}          & \multicolumn{1}{c|}{0.5242} & 0.5459               \\
\multicolumn{1}{c|}{}                        & \multicolumn{1}{l|}{Search QA}             & \multicolumn{1}{c|}{0.7946} & 0.814                \\
\multicolumn{1}{c|}{}                        & \multicolumn{1}{l|}{Title Generation}      & \multicolumn{1}{c|}{0.535} & 0.5729               \\ \hline
\multicolumn{1}{c|}{\multirow{4}{*}{\shortstack{Standard \\ (Non-repetition)}}}     & \multicolumn{1}{l|}{Sentiment Analysis}      & \multicolumn{1}{c|}{0.9553} & 0.9658               \\
\multicolumn{1}{c|}{}                        & \multicolumn{1}{l|}{Knowledge QA}          & \multicolumn{1}{c|}{0.4911} & 0.4823               \\
\multicolumn{1}{c|}{}                        & \multicolumn{1}{l|}{Search QA}             & \multicolumn{1}{c|}{0.8208} & 0.838               \\
\multicolumn{1}{c|}{}                        & \multicolumn{1}{l|}{Title Generation}      & \multicolumn{1}{c|}{0.6618} & 0.6815               \\ \hline
\end{tabular}%
}
\caption{Comparison between the results of few-shot with arbitrary format and standard format. The details can be found in Table \ref{tab:tab101} and Table \ref{tab:tab107}.}
\label{tab:tab7}
\end{table}

\section{Comparison with Other Method}

\subsection{Comparison with Other Automatic Evaluation Method}

We list Spearman correlation of 3 typical method and our method here: GPTScore (0.1888), TigerScore (0.3373), MT-Bench (0.6/0.85) and TALEC (0.8962/0.875). It is difficult to make a completely fair side-by-side comparison with other methods due to the differences in the score system, evaluation question types, and evaluation criteria. However, the high alignment of TALEC compared to other methods can also indirectly indicate its validity and usability.

\subsection{Comparison with Human Annotation}

Compared to automated evaluation, human assessment possesses a greater capacity to encompass the intricate and adaptable evaluation criteria and scales inherent in our business operations. Nevertheless, the financial implications of employing human evaluators are substantial, with the primary expenditure being personnel training costs. This is particularly true for specialized domains where the evaluator requirements are notably stringent. Furthermore, the efficiency of human assessment significantly lags behind that of automated evaluation, thereby considerably impeding the iterative development of large language models.

Additionally, human annotation, while useful, is not always dependable. In the initial phases of our experiment, we utilized manual evaluation to gather enough data for the development of a more effective assessment system. This process involved a dual-review system, blind reviews, comprehensive quality checks for contentious cases, random spot checks for non-contentious cases, and a concluding round of spot checks. Only through the application of these multiple strategies and checks were we able to ensure the data's accuracy. However, an analysis of the manual annotation results prior to inspection revealed a significant lack of alignment among different annotators in the absence of rigorous quality control, as demonstrated in Table \ref{tab:tab8}. This discrepancy can be partially attributed to the complexity of the custom business evaluation criteria. Despite these challenges, automated evaluation has proven to be highly effective in such a demanding context.

\begin{table}[htbp]
\centering
\scalebox{1.0}{%
\begin{tabular}{@{}l|l}
\toprule
\textbf{Task}             & \textbf{Spearman}                                                                                          \\ \midrule
Sentiment Analysis                  & 0.9054                                                              \\
Knowledge QA    & 0.6523                                                                 \\
Search QA               & 0.7973                                       \\
Title Generation               & 0.8772                                   
                                  \\ \bottomrule
\end{tabular}%
}
\caption{Alignment degree of the manual annotations before quality inspection.}
\label{tab:tab8}
\end{table}

\section{Conclusion}

We propose a method: TALEC, which allows users to flexibly set their own evaluation criteria, and uses in-context learning (ICL) to teach judge model these in-house criteria. We try many approach to improve the judge abilities of model, such as criteria division and combining zero-shot with few-shot. We also come up with an engineering approach to adjust and iterate shots. which splits the dataset and simulates the practice-quality inspection cycle process. In addition, we find that when injecting manually-written shots into the context of a model, the model will not only try to understand the shots but also imitate the writing style of shots. This imitation may make the model ignore some key information and drop its CoT ability. We then compare fine-tuning with ICL, finding that fine-tuning can be replaced by ICL. In the end, we compare TALEC with other methods and humans, verifying the availability of TALEC.

\section{Limitations}

Although TALEC outperforms than many other methods, it still makes some really stupid mistakes sometimes. And TALEC relies heavily on manual annotation in the early stage, making it hard to start. Some other methods also focus on special abilities like hallucination and contextual memory, but TALEC can't evaluate these abilities so far.

\bibliography{custom}

\appendix

\section{Prompt Template}

\begin{figure*}[htbp]
    \centering
    \includegraphics[width=1.0\linewidth]{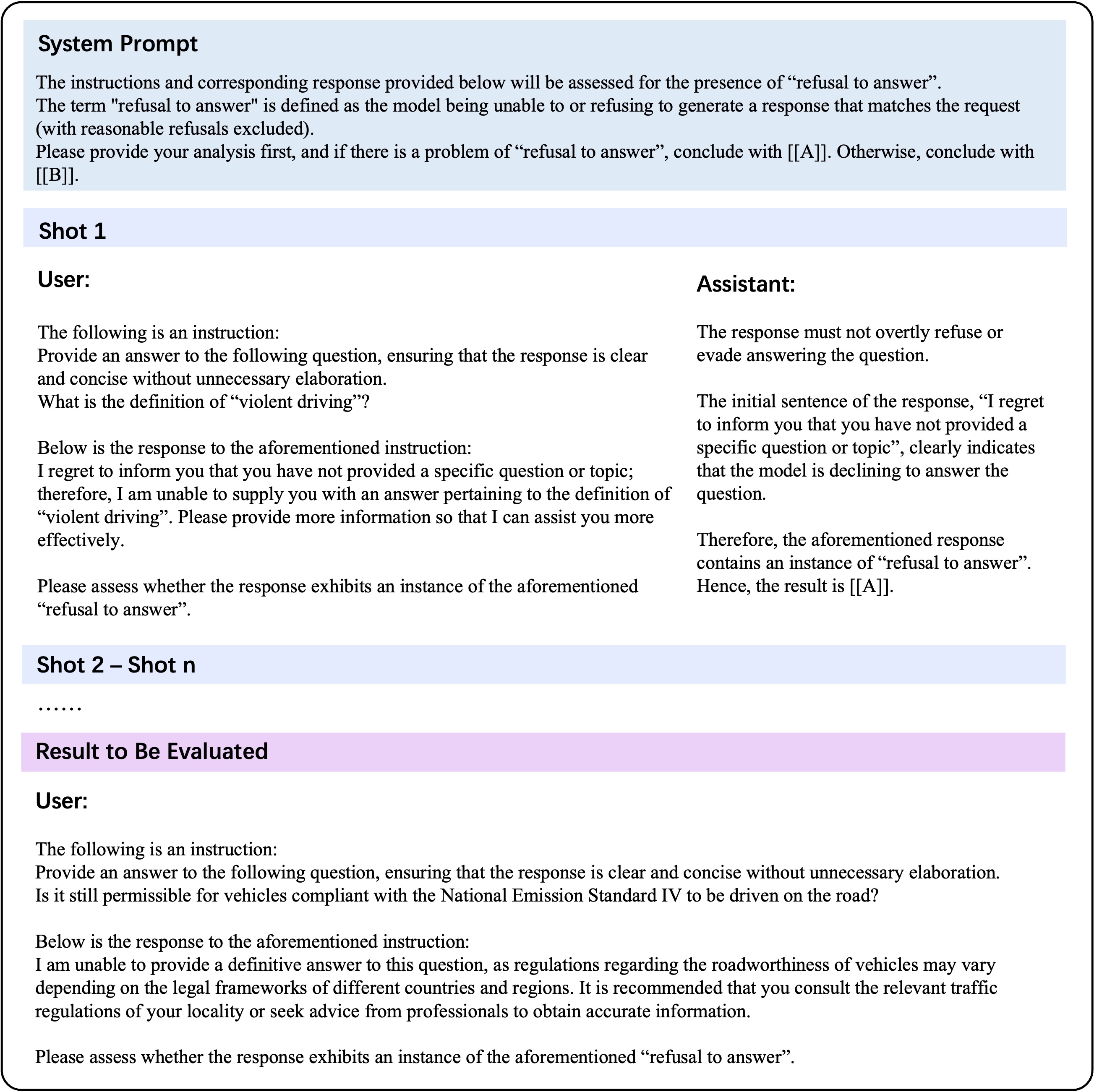}
    \caption{The prompt format for evaluation tasks is depicted in the following sequence: the system-generated prompt, containing explicit task requirements and evaluation criteria explanations, aids the model in comprehending the assessment's scope and complexity. This is succeeded by the few-shot section, which presents a variety of examples in the same format, including both positive and negative instances, enabling the model to adopt the sample structure for its output. Finally, the output produced by LLM, is the subject of evaluation.}
    \label{fig:figure-3}
\end{figure*}

\begin{figure*}[htbp]
    \centering
    \includegraphics[width=1.0\linewidth]{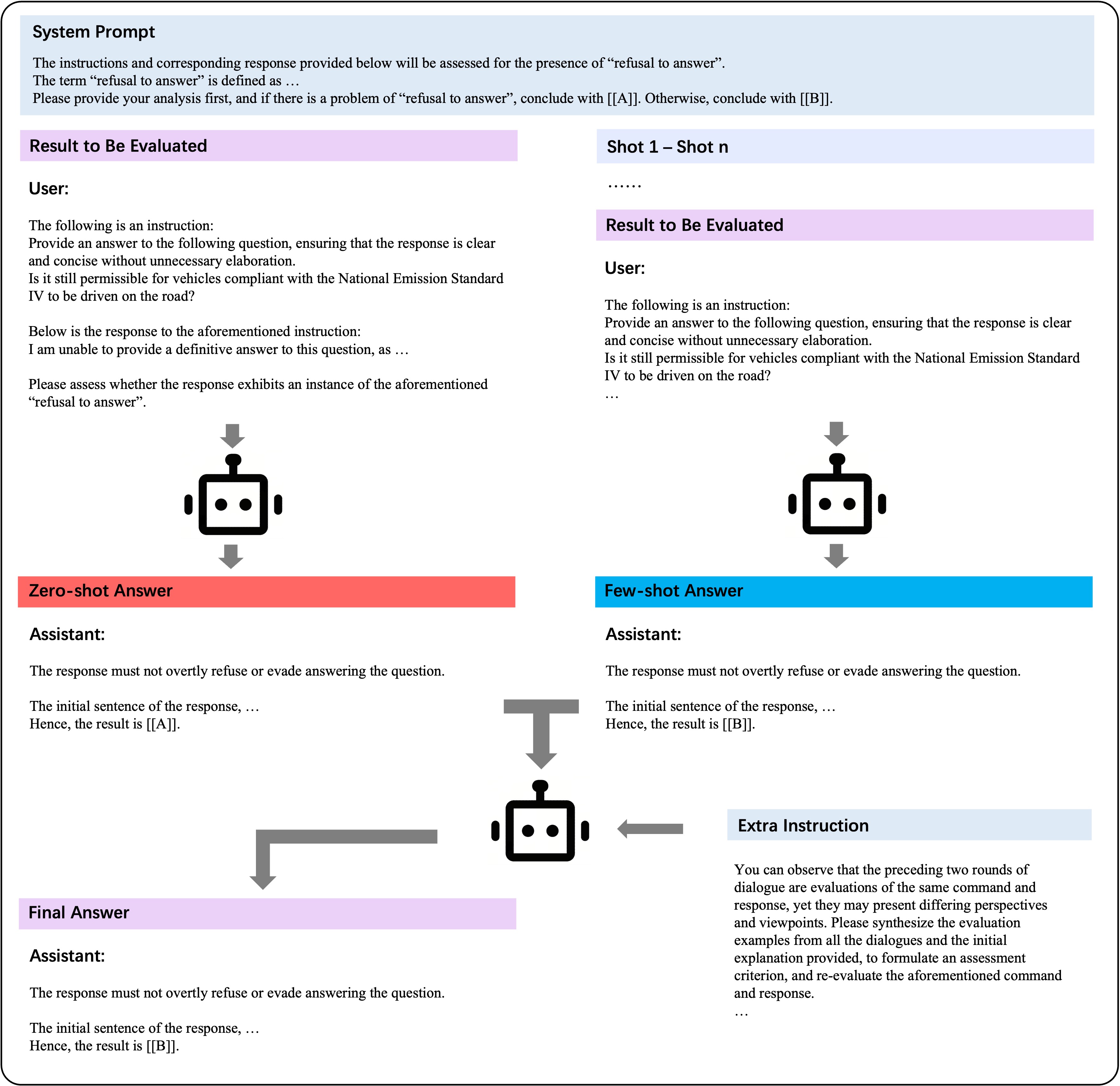}
    \caption{This figure illustrates an instance of the prompt format for Combining Zero-shot with Few-shot. We combine system Prompt, Shots, answer output by zero-shot, answer output by few-shot to get a new context and use this con-text to judge again. Then we get the final result. }
    \label{fig:figure-4}
\end{figure*}

The prompt templates are exemplified in the Figure \ref{fig:figure-3} and Figure \ref{fig:figure-4}.

\section{Experimental Results}

The details of all experimental results are tabulated in Table \ref{tab:tab101} to Table \ref{tab:tab107}.

\begin{table*}
\centering
\scalebox{0.5}{
\begin{tabular}{|c|c|c|c|c|c|c|c|c|}
\hline \multirow{2}{*}{ Method } & \multirow{2}{*}{ task } & \multirow{2}{*}{ label } & \multicolumn{2}{|c|}{ Spearman/Pearson/Kendall } & \multicolumn{2}{|c|}{ Acc } & \multicolumn{2}{|c|}{ F1/Precision/Recall } \\
\cline{4-9} & & & eval & test & eval & test & eval & test \\
\hline \multirow{31}{*}{ Arbitrariness } & \multirow{8}{*}{\begin{tabular}{l} 
Sentiment Analysis
\end{tabular}} & Not Meeting the Requirements & \multirow{8}{*}{\begin{tabular}{l}
\begin{tabular}{l}
0.9551 \\
0.9523 \\
0.9484
\end{tabular}
\end{tabular}} & \multirow{8}{*}{\begin{tabular}{l}
\begin{tabular}{l}
0.9515 \\
0.943 \\
0.9268
\end{tabular}
\end{tabular}} & 0.9123 & 0.8829 &  0.75 / 0.625 / 0.9375  &  0.6286 / 0.4783 / 0.9167  \\
\cline{3-3} \cline{6-9} & & Incorrect Answer/Unrelated Matching Results & & & 0.9737 & 0.9279 &  0.9684 / 1.0 / 0.9388  &  0.92 / 0.9388 / 0.902  \\
\cline{3-3} \cline{6-9} & &  Refusal to Answer   & & & 0.9737 & 0.964 &  0.6667 / 0.75 / 0.6  &  0.5 / 0.3333 / 1.0  \\
\cline{3-3} \cline{6-9} & & Untranslated Text & & & 0.9825 & 0.982 &  0.875 / 0.7778 / 1.0  &  0.875 / 0.7778 / 1.0  \\
\cline{3-3} \cline{6-9} & &  Confusing Answers  & & & 0.9912 & 0.982 &  0.8571 / 0.75 / 1.0  &  0.75 / 0.6 / 1.0  \\
\cline{3-3} \cline{6-9} & & Stiffness & & & 0.9649 & 0.9279 &  0.8182 / 0.6923 / 1.0  &  0.6667 / 0.5 / 1.0  \\
\cline{3-3} \cline{6-9} & &  Repetitive Expression  & & & 0.9649 & 0.955 &  0.7143 / 1.0 / 0.5556  &  0.5455 / 1.0 / 0.375  \\
\cline{3-3} \cline{6-9} & &  Subject Imprecision  & & & 0.8947 & 0.7838 &  0.7857 / 0.7333 / 0.8462  &  0.3333 / 0.2069 / 0.8571  \\
\cline{2-9} & \multirow{8}{*}{ Knowledge QA } &  Incorrect Answer/Unrelated Matching Results  & \multirow{8}{*}{\begin{tabular}{l}
\begin{tabular}{l}
0.5242 \\
0.5345 \\
0.4974
\end{tabular}
\end{tabular}} & \multirow{8}{*}{\begin{tabular}{l}
\begin{tabular}{l}
0.5459 \\
0.5436 \\
0.518
\end{tabular}
\end{tabular}} & 0.8085 & 0.7591 &  0.4906 / 0.3333 / 0.9286  &  0.2979 / 0.1944 / 0.6364  \\
\cline{3-3} \cline{6-9} & &  Refusal to Answer   & & & 1 & 0.9927 & 1.0/1.0/1.0 &  0.8571 / 0.75 / 1.0  \\
\cline{3-3} \cline{6-9} & &  Untranslated Text  & & & 0.9858 & 0.9927 &  0.75 / 0.75 / 0.75  &  0.8571 / 1.0 / 0.75  \\
\cline{3-3} \cline{6-9} & &  Confusing Answers  & & & 0.9858 & 1 &  0.75 / 0.6 / 1.0  & 1.0/1.0/1.0 \\
\cline{3-3} \cline{6-9} & & Incomplete Answers & & & 0.7589 & 0.6934 &  0.6136 / 0.7941 / 0.5  &  0.5625 / 0.871 / 0.4154  \\
\cline{3-3} \cline{6-9} & &  Stiffness  & & & 0.9716 & 1 &  0.7143 / 0.8333 / 0.625  & 1.0/1.0/1.0 \\
\cline{3-3} \cline{6-9} & & Repetitive Expression & & & 0.9858 & 0.9781 &  0.8571 / 0.8571 / 0.8571  &  0.7273 / 0.8 / 0.6667  \\
\cline{3-3} \cline{6-9} & & Subject Imprecision & & & 0.9645 & 0.9416 &  0.7619 / 0.8889 / 0.6667  &  0.3333 / 0.4 / 0.2857  \\
\cline{2-9} & \multirow{7}{*}{ Search QA } & Incorrect Answer/Unrelated Matching Results & \multirow{7}{*}{\begin{tabular}{l}
\begin{tabular}{l}
0.7946 \\
0.7901 \\
0.7852
\end{tabular}
\end{tabular}} & \multirow{7}{*}{\begin{tabular}{l}
\begin{tabular}{l}
0.814 \\
0.818 \\
0.791
\end{tabular}
\end{tabular}} & 0.9449 & 0.9587 &  0.7586 / 0.7857 / 0.7333  &  0.8387 / 0.9286 / 0.7647  \\
\cline{3-3} \cline{6-9} & & External Links or Diversions & & & 0.9921 & 0.9752 &  0.9333 / 1.0 / 0.875  &  0.7273 / 0.6667 / 0.8  \\
\cline{3-3} \cline{6-9} & & Refusal to Answer & & & 0.9685 & 1 &  0.7143 / 0.5556 / 1.0  &  1.0 / 1.0 / 1.0  \\
\cline{3-3} \cline{6-9} & &  Incomplete Answers  & & & 0.8661 & 0.9174 &  0.32 / 0.3636 / 0.2857  &  0.5455 / 0.8571 / 0.4  \\
\cline{3-3} \cline{6-9} & & Stiffness & & & 0.9606 & 0.9752 &  0.4444 / 0.2857 / 1.0  &  0.6667 / 0.75 / 0.6  \\
\cline{3-3} \cline{6-9} & & Repetitive Expression & & & 0.9528 & 0.9421 &  0.4 / 0.2857 / 0.6667  &  0.3636 / 0.2222 / 1.0  \\
\cline{3-3} \cline{6-9} & & Subject Imprecision & & & 0.9528 & 0.9174 &  0.5714 / 0.4 / 1.0  &  0.5455 / 0.6 / 0.5  \\
\cline{2-9} & \multirow{8}{*}{ Title Generation } & Not Meeting the Requirements & \multirow{8}{*}{\begin{tabular}{l}
\begin{tabular}{l}
0.535 \\
0.5629 \\
0.5153
\end{tabular}
\end{tabular}} & \multirow{8}{*}{\begin{tabular}{l}
\begin{tabular}{l}
0.5729 \\
0.593 \\
0.5483
\end{tabular}
\end{tabular}} & 0.617 & 0.6147 &  0.6457 / 0.7387 / 0.5734  &  0.6426 / 0.7477 / 0.5634  \\
\cline{3-3} \cline{6-9} & & Incorrect Answer/Unrelated Matching Results & & & 0.9064 & 0.8745 &  0.8036 / 0.7377 / 0.8824  &  0.7752 / 0.7937 / 0.7576  \\
\cline{3-3} \cline{6-9} & & External Links or Diversion & & & 1 & 1 & 1.0/1.0/1.0 & 1.0/1.0/1.0 \\
\cline{3-3} \cline{6-9} & &  Untranslated Text & & & 0.9745 & 0.987 &  0.625 / 0.5 / 0.8333  &  0.7273 / 1.0 / 0.5714  \\
\cline{3-3} \cline{6-9} & & Confusing Answers & & & 0.966 & 0.9827 &  0.6923 / 0.5625 / 0.9  &  0.7778 / 0.6364 / 1.0  \\
\cline{3-3} \cline{6-9} & & Stiffness & & & 0.9787 & 0.9481 &  0.7368 / 0.6364 / 0.875  &  0.6842 / 1.0 / 0.52  \\
\cline{3-3} \cline{6-9} & & Repetitive Expression & & & 0.9745 & 0.961 &  0.75 / 0.6 / 1.0  &  0.5263 / 0.3571 / 1.0  \\
\cline{3-3} \cline{6-9} & & Subject Imprecision & & & 0.9277 & 0.9394 &  0.7213 / 0.7857 / 0.6667  &  0.72 / 0.8182 / 0.6429  \\
\hline
\end{tabular}
}
\caption{The overall results of Arbitrariness.}
\label{tab:tab101}
\end{table*}

\begin{table*}
\centering
\scalebox{0.5}{
\begin{tabular}{|c|c|c|c|c|c|c|c|c|}
\hline \multirow{2}{*}{ Method } & \multirow{2}{*}{ task } & \multirow{2}{*}{ label } & \multicolumn{2}{|c|}{ Spearman/Pearson/Kendall } & \multicolumn{2}{|c|}{ Acc } & \multicolumn{2}{|c|}{ F1/Precision/Recall } \\
\cline{4-9} & & & eval & test & eval & test & eval & test \\
\hline \multirow{31}{*}{\shortstack{Standard Prompt \\ Paradigm} } & \multirow{8}{*}{\begin{tabular}{l} 
Sentiment Analysis
\end{tabular}} & Not Meeting the Requirements & \multirow{8}{*}{\begin{tabular}{l}
\begin{tabular}{l}
0.9579 \\
0.9564 \\
0.9545
\end{tabular}
\end{tabular}} & \multirow{8}{*}{\begin{tabular}{l}
\begin{tabular}{l}
0.9693 \\
0.9643 \\
0.9584
\end{tabular}
\end{tabular}} & 0.9649 & 0.9279 &  0.875 / 0.875 / 0.875  &  0.7143 / 0.625 / 0.8333  \\
\cline{3-3} \cline{6-9} & & Incorrect Answer/Unrelated Matching Results & & & 0.9825 & 0.9189 &  0.9792 / 1.0 / 0.9592  &  0.9091 / 0.9375 / 0.8824  \\
\cline{3-3} \cline{6-9} & &  Refusal to Answer   & & & 0.9825 & 0.982 & 0.75/1.0/0.6 &  0.6667 / 0.5 / 1.0  \\
\cline{3-3} \cline{6-9} & & Untranslated Text & & & 1 & 1 & 1.0/1.0/1.0 & 1.0/1.0/1.0  \\
\cline{3-3} \cline{6-9} & &  Confusing Answers  & & & 0.9912 & 0.982 &  0.8 / 1.0 / 0.6667  &  0.6667 / 0.6667 / 0.6667  \\
\cline{3-3} \cline{6-9} & & Stiffness & & & 0.9825 & 0.964 &  0.875 / 1.0 / 0.7778  & 0.8/0.6667/1.0  \\
\cline{3-3} \cline{6-9} & &  Repetitive Expression  & & & 0.9912 & 0.9369 &  0.9412 / 1.0 / 0.8889  &  0.6667 / 0.5385 / 0.875  \\
\cline{3-3} \cline{6-9} & &  Subject Imprecision  & & & 0.9123 & 0.8829 &  0.7727 / 0.9444 / 0.6538  &  0.48 / 0.3333 / 0.8571  \\
\cline{2-9} & \multirow{8}{*}{ Knowledge QA } &  Incorrect Answer/Unrelated Matching Results  & \multirow{8}{*}{\begin{tabular}{l}
\begin{tabular}{l}
0.4945 \\
0.5153 \\
0.4734
\end{tabular}
\end{tabular}} & \multirow{8}{*}{\begin{tabular}{l}
\begin{tabular}{l}
0.5063 \\
0.4982 \\
0.48
\end{tabular}
\end{tabular}} & 0.8298 & 0.7737 &  0.52 / 0.3611 / 0.9286  &  0.2439 / 0.1667 / 0.4545  \\
\cline{3-3} \cline{6-9} & &  Refusal to Answer   & & & 1 & 1 & 1.0/1.0/1.0 & 1.0/1.0/1.0  \\
\cline{3-3} \cline{6-9} & &  Untranslated Text  & & & 0.9929 & 0.9927 & 0.8889/0.8/1.0 &  0.8571 / 1.0 / 0.75  \\
\cline{3-3} \cline{6-9} & &  Confusing Answers  & & & 0.9787 & 1 &  0.6667 / 0.5 / 1.0  & 1.0/1.0/1.0 \\
\cline{3-3} \cline{6-9} & & Incomplete Answers & & & 0.7589 & 0.7372 &  0.575 / 0.8846 / 0.4259  &  0.625 / 0.9677 / 0.4615  \\
\cline{3-3} \cline{6-9} & &  Stiffness  & & & 0.9645 & 1 &  0.6154 / 0.8 / 0.5  & 1.0/1.0/1.0 \\
\cline{3-3} \cline{6-9} & & Repetitive Expression & & & 0.9929 & 0.9781 &  0.9333 / 0.875 / 1.0  &  0.7692 / 0.7143 / 0.8333   \\
\cline{3-3} \cline{6-9} & & Subject Imprecision & & & 0.9574 & 0.9562 &  0.7273 / 0.8 / 0.6667  & 0.5/0.6/0.4286  \\
\cline{2-9} & \multirow{7}{*}{ Search QA } & Incorrect Answer/Unrelated Matching Results & \multirow{7}{*}{\begin{tabular}{l}
\begin{tabular}{l}
0.8263 \\
0.8223 \\
0.7951
\end{tabular}
\end{tabular}} & \multirow{7}{*}{\begin{tabular}{l}
\begin{tabular}{l}
0.8487 \\
0.8487 \\
0.8352
\end{tabular}
\end{tabular}} & 0.9528 & 0.9421 &  0.7692 / 0.9091 / 0.6667  &  0.7586 / 0.9167 / 0.6471  \\
\cline{3-3} \cline{6-9} & & External Links or Diversions & & & 1 & 0.9835 & 1.0/1.0/1.0 &  0.8333 / 0.7143 / 1.0  \\
\cline{3-3} \cline{6-9} & & Refusal to Answer & & & 0.9685 & 1 &  0.7143 / 0.5556 / 1.0  & 1.0/1.0/1.0  \\
\cline{3-3} \cline{6-9} & &  Incomplete Answers  & & & 0.8819 & 0.9091 &  0.4444 / 0.4615 / 0.4286  &  0.4211 / 1.0 / 0.2667  \\
\cline{3-3} \cline{6-9} & & Stiffness & & & 0.9606 & 0.9835 &  0.4444 / 0.2857 / 1.0  & 0.75/1.0/0.6  \\
\cline{3-3} \cline{6-9} & & Repetitive Expression & & & 0.9843 & 0.9917 &  0.6667 / 0.6667 / 0.6667  & 0.8/0.6667/1.0  \\
\cline{3-3} \cline{6-9} & & Subject Imprecision & & & 0.9449 & 0.9256 &  0.3636 / 0.2857 / 0.5  & 0.4706/0.8/0.3333  \\
\cline{2-9} & \multirow{8}{*}{ Title Generation } & Not Meeting the Requirements & \multirow{8}{*}{\begin{tabular}{l}
\begin{tabular}{l}
0.9207 \\
0.9322 \\
0.915
\end{tabular}
\end{tabular}} & \multirow{8}{*}{\begin{tabular}{l}
\begin{tabular}{l}
0.9006 \\
0.9055 \\
0.8963
\end{tabular}
\end{tabular}} & 0.9574 & 0.9524 &  0.9695 / 0.9636 / 0.9755  &  0.9657 / 0.9873 / 0.9451  \\
\cline{3-3} \cline{6-9} & & Incorrect Answer/Unrelated Matching Results & & & 0.9319 & 0.8788 &  0.8298 / 0.907 / 0.7647  &  0.7544 / 0.8958 / 0.6515  \\
\cline{3-3} \cline{6-9} & & External Links or Diversion & & & 1 & 1 &  1.0 / 1.0 / 1.0  &  1.0 / 1.0 / 1.0 \\
\cline{3-3} \cline{6-9} & &  Untranslated Text & & & 0.9745 & 1 &  0.6667 / 0.5 / 1.0  &  1.0 / 1.0 / 1.0   \\
\cline{3-3} \cline{6-9} & & Confusing Answers & & & 0.983 & 0.9913 &  0.8 / 0.8 / 0.8  &  0.8571 / 0.8571 / 0.8571   \\
\cline{3-3} \cline{6-9} & & Stiffness & & & 0.9404 & 0.9827 &  0.5333 / 0.3636 / 1.0  &  0.913 / 1.0 / 0.84  \\
\cline{3-3} \cline{6-9} & & Repetitive Expression & & & 0.9745 & 0.9913 &  0.6667 / 0.6667 / 0.6667  &  0.8333 / 0.7143 / 1.0  \\
\cline{3-3} \cline{6-9} & & Subject Imprecision & & & 0.9277 & 0.9567 &  0.7213 / 0.7857 / 0.6667  & 0.7917/0.95/0.6786  \\
\hline
\end{tabular}
}
\caption{The overall results of Standard Prompt Paradigm}
\label{tab:tab102}
\end{table*}

\begin{table*}
\centering
\scalebox{0.5}{
\begin{tabular}{|c|c|c|c|c|c|c|c|c|}
\hline \multirow{2}{*}{ Method } & \multirow{2}{*}{ task } & \multirow{2}{*}{ label } & \multicolumn{2}{|c|}{ Spearman/Pearson/Kendall } & \multicolumn{2}{|c|}{ Acc } & \multicolumn{2}{|c|}{ F1/Precision/Recall } \\
\cline{4-9} & & & eval & test & eval & test & eval & test \\
\hline \multirow{31}{*}{\shortstack{Multi-turn \\ with Zero-shot}} & \multirow{8}{*}{\begin{tabular}{l} 
Sentiment Analysis
\end{tabular}} & Not Meeting the Requirements & \multirow{8}{*}{\begin{tabular}{l}
\begin{tabular}{l}
0.9536 \\
0.9506 \\
0.9467
\end{tabular}
\end{tabular}} & \multirow{8}{*}{\begin{tabular}{l}
\begin{tabular}{l}
0.9485 \\
0.943 \\
0.9378
\end{tabular}
\end{tabular}} & 0.9825 & 0.8919 &  0.9412 / 0.8889 / 1.0  &  0.6471 / 0.5 / 0.9167  \\
\cline{3-3} \cline{6-9} & & Incorrect Answer/Unrelated Matching Results & & & 0.9825 & 0.9369 &  0.9792 / 1.0 / 0.9592  &  0.932 / 0.9231 / 0.9412  \\
\cline{3-3} \cline{6-9} & &  Refusal to Answer   & & & 0.9737 & 0.9459 &  0.7692 / 0.625 / 1.0  &  0.4 / 0.25 / 1.0  \\
\cline{3-3} \cline{6-9} & & Untranslated Text & & & 0.9912 & 0.991 &  0.9333 / 0.875 / 1.0  &  0.9231 / 1.0 / 0.8571  \\
\cline{3-3} \cline{6-9} & &  Confusing Answers  & & & 0.9912 & 0.991 & 0.8/1.0/0.6667 &  0.8 / 1.0 / 0.6667  \\
\cline{3-3} \cline{6-9} & & Stiffness & & & 0.9825 & 0.964 &  0.9 / 0.8182 / 1.0  & 0.8/0.6667/1.0  \\
\cline{3-3} \cline{6-9} & &  Repetitive Expression  & & & 0.9912 & 0.955 &  0.9412 / 1.0 / 0.8889  &  0.7368 / 0.6364 / 0.875  \\
\cline{3-3} \cline{6-9} & &  Subject Imprecision  & & & 0.9825 & 0.8649 &  0.963 / 0.9286 / 1.0  &  0.4444 / 0.3 / 0.8571  \\
\cline{2-9} & \multirow{8}{*}{ Knowledge QA } &  Incorrect Answer/Unrelated Matching Results  & \multirow{8}{*}{\begin{tabular}{l}
\begin{tabular}{l}
0.8597 \\
0.8542 \\
0.8487
\end{tabular}
\end{tabular}} & \multirow{8}{*}{\begin{tabular}{l}
\begin{tabular}{l}
0.7089 \\
0.6794 \\
0.6898
\end{tabular}
\end{tabular}} & 0.9645 & 0.9051 &  0.8148 / 0.8462 / 0.7857  &  0.5185 / 0.4375 / 0.6364  \\
\cline{3-3} \cline{6-9} & &  Refusal to Answer   & & & 0.9929 & 1 & 0.8/1.0/0.6667 & 1.0/1.0/1.0  \\
\cline{3-3} \cline{6-9} & &  Untranslated Text  & & & 0.9929 & 1 & 0.8889/0.8/1.0 & 1.0/1.0/1.0  \\
\cline{3-3} \cline{6-9} & &  Confusing Answers  & & & 0.9929 & 0.9927 &  0.8 / 1.0 / 0.6667  & 0.8889/1.0/0.8 \\
\cline{3-3} \cline{6-9} & & Incomplete Answers & & & 0.9433 & 0.9197 &  0.9245 / 0.9423 / 0.9074  &  0.912 / 0.95 / 0.8769  \\
\cline{3-3} \cline{6-9} & &  Stiffness  & & & 0.9787 & 0.9854 &  0.8 / 0.8571 / 0.75  &  0.875 / 0.7778 / 1.0 \\
\cline{3-3} \cline{6-9} & & Repetitive Expression & & & 1 & 0.9854 & 1.0/1.0/1.0 &  0.8333 / 0.8333 / 0.8333   \\
\cline{3-3} \cline{6-9} & & Subject Imprecision & & & 0.9716 & 0.9635 &  0.8182 / 0.9 / 0.75  &  0.6154 / 0.6667 / 0.5714  \\
\cline{2-9} & \multirow{7}{*}{ Search QA } & Incorrect Answer/Unrelated Matching Results & \multirow{7}{*}{\begin{tabular}{l}
\begin{tabular}{l}
0.8574 \\
0.8566 \\
0.8377
\end{tabular}
\end{tabular}} & \multirow{7}{*}{\begin{tabular}{l}
\begin{tabular}{l}
0.9438 \\
0.9434 \\
0.9394
\end{tabular}
\end{tabular}} & 0.9449 & 0.9835 &  0.7586 / 0.7857 / 0.7333  &  0.9412 / 0.9412 / 0.9412  \\
\cline{3-3} \cline{6-9} & & External Links or Diversions & & & 1 & 1 &  1.0 / 1.0 / 1.0  &  1.0 / 1.0 / 1.0  \\
\cline{3-3} \cline{6-9} & & Refusal to Answer & & & 0.9843 & 1 &  0.8333 / 0.7143 / 1.0  & 1.0/1.0/1.0  \\
\cline{3-3} \cline{6-9} & &  Incomplete Answers  & & & 0.9213 & 0.9917 &  0.7222 / 0.5909 / 0.9286  &  0.9677 / 0.9375 / 1.0  \\
\cline{3-3} \cline{6-9} & & Stiffness & & & 0.9685 & 0.9917 & 0.5/0.3333/1.0 & 0.8889/1.0/0.8  \\
\cline{3-3} \cline{6-9} & & Repetitive Expression & & & 0.9843 & 0.9917 &  0.6667 / 0.6667 / 0.6667  &  0.6667 / 1.0 / 0.5  \\
\cline{3-3} \cline{6-9} & & Subject Imprecision & & & 0.9528 & 0.9835 &  0.5714 / 0.4 / 1.0  &  0.9231 / 0.8571 / 1.0  \\
\cline{2-9} & \multirow{8}{*}{ Title Generation } & Not Meeting the Requirements & \multirow{8}{*}{\begin{tabular}{l}
\begin{tabular}{l}
0.915 \\
0.9142 \\
0.9124
\end{tabular}
\end{tabular}} & \multirow{8}{*}{\begin{tabular}{l}
\begin{tabular}{l}
0.9206 \\
0.9243 \\
0.9148
\end{tabular}
\end{tabular}} & 0.983 & 0.9697 &  0.9878 / 0.9818 / 0.9939  &  0.9785 / 0.9876 / 0.9695  \\
\cline{3-3} \cline{6-9} & & Incorrect Answer/Unrelated Matching Results & & & 0.8681 & 0.9957 &  0.7597 / 0.6282 / 0.9608  &  0.9925 / 0.9851 / 1.0   \\
\cline{3-3} \cline{6-9} & & External Links or Diversion & & & 1 & 1 &  1.0 / 1.0 / 1.0  &  1.0 / 1.0 / 1.0 \\
\cline{3-3} \cline{6-9} & &  Untranslated Text & & & 0.983 & 1 &  0.75 / 0.6 / 1.0  & 1.0/1.0/1.0   \\
\cline{3-3} \cline{6-9} & & Confusing Answers & & & 1 & 1 &  1.0 / 1.0 / 1.0  &  1.0 / 1.0 / 1.0    \\
\cline{3-3} \cline{6-9} & & Stiffness & & & 0.9872 & 1 &  0.8421 / 0.7273 / 1.0  & 1.0/1.0/1.0  \\
\cline{3-3} \cline{6-9} & & Repetitive Expression & & & 0.9957 & 1 &  0.9474 / 0.9 / 1.0  & 1.0/1.0/1.0  \\
\cline{3-3} \cline{6-9} & & Subject Imprecision & & & 0.9489 & 0.9957 &  0.8462 / 0.7333 / 1.0  &  0.9825 / 0.9655 / 1.0  \\
\hline
\end{tabular}
}
\caption{The overall results of Multi-turn with Zero-shot}
\label{tab:tab103}
\end{table*}

\begin{table*}
\centering
\scalebox{0.5}{
\begin{tabular}{|c|c|c|c|c|c|c|c|c|}
\hline \multirow{2}{*}{ Method } & \multirow{2}{*}{ task } & \multirow{2}{*}{ label } & \multicolumn{2}{|c|}{ Spearman/Pearson/Kendall } & \multicolumn{2}{|c|}{ Acc } & \multicolumn{2}{|c|}{ F1/Precision/Recall } \\
\cline{4-9} & & & eval & test & eval & test & eval & test \\
\hline \multirow{31}{*}{ \shortstack{Qwen-72B-Chat \\ + SFT} } & \multirow{8}{*}{\begin{tabular}{l} 
Sentiment Analysis
\end{tabular}} & Not Meeting the Requirements & \multirow{8}{*}{\begin{tabular}{l}
\begin{tabular}{l}
0.4565 \\
0.4166 \\
0.4366
\end{tabular}
\end{tabular}} & \multirow{8}{*}{\begin{tabular}{l}
\begin{tabular}{l}
0.4211 \\
0.4308 \\
0.4025
\end{tabular}
\end{tabular}} & 0.8158 & 0.7297 &  0.16 / 0.2222 / 0.125  &  0.2105 / 0.1538 / 0.3333  \\
\cline{3-3} \cline{6-9} & & Incorrect Answer/Unrelated Matching Results & & & 0.8158 & 0.8108 &  0.7586 / 0.8684 / 0.6735  &  0.764 / 0.8947 / 0.6667  \\
\cline{3-3} \cline{6-9} & &  Refusal to Answer   & & & 0.9386 & 0.9459 &  0.5333 / 0.4 / 0.8  &  0.4 / 0.25 / 1.0  \\
\cline{3-3} \cline{6-9} & & Untranslated Text & & & 0.8772 & 0.7928 &  0.4167 / 0.2941 / 0.7143  &  0.303 / 0.1923 / 0.7143  \\
\cline{3-3} \cline{6-9} & &  Confusing Answers  & & & 0.8596 & 0.8468 &  0.2 / 0.1176 / 0.6667  &  0.1905 / 0.1111 / 0.6667   \\
\cline{3-3} \cline{6-9} & & Stiffness & & & 0.5702 & 0.4865 &  0.1967 / 0.1154 / 0.6667  &  0.1739 / 0.0984 / 0.75  \\
\cline{3-3} \cline{6-9} & &  Repetitive Expression  & & & 0.4737 & 0.4865 &  0.1892 / 0.1077 / 0.7778  &  0.1972 / 0.1111 / 0.875  \\
\cline{3-3} \cline{6-9} & &  Subject Imprecision  & & & 0.5877 & 0.6216 &  0.3188 / 0.2558 / 0.4231  &  0.087 / 0.0513 / 0.2857   \\
\cline{2-9} & \multirow{8}{*}{ Knowledge QA } &  Incorrect Answer/Unrelated Matching Results  & \multirow{8}{*}{\begin{tabular}{l}
\begin{tabular}{l}
0.1886 \\
0.1896 \\
0.1742
\end{tabular}
\end{tabular}} & \multirow{8}{*}{\begin{tabular}{l}
\begin{tabular}{l}
0.2444 \\
0.2142 \\
0.2301
\end{tabular}
\end{tabular}} & 0.7305 & 0.5766 &  0.3214 / 0.2143 / 0.6429  &  0.1471 / 0.0877 / 0.4545  \\
\cline{3-3} \cline{6-9} & &  Refusal to Answer   & & & 0.922 & 0.927 &  0.3529 / 0.2143 / 1.0  &  0.375 / 0.2308 / 1.0  \\
\cline{3-3} \cline{6-9} & &  Untranslated Text  & & & 0.7305 & 0.7737 &  0.0952 / 0.0526 / 0.5  &  0.1143 / 0.0645 / 0.5  \\
\cline{3-3} \cline{6-9} & &  Confusing Answers  & & & 0.8865 & 0.8248 &  0.2 / 0.1176 / 0.6667  &  0.2 / 0.12 / 0.6 \\
\cline{3-3} \cline{6-9} & & Incomplete Answers & & & 0.5887 & 0.5985 &  0.4528 / 0.4615 / 0.4444  &  0.56 / 0.5833 / 0.5385  \\
\cline{3-3} \cline{6-9} & &  Stiffness  & & & 0.617 & 0.6058 &  0.1562 / 0.0893 / 0.625  &  0.1818 / 0.1017 / 0.8571 \\
\cline{3-3} \cline{6-9} & & Repetitive Expression & & & 0.4468 & 0.438 &  0.1333 / 0.0723 / 0.8571  &  0.1348 / 0.0723 / 1.0   \\
\cline{3-3} \cline{6-9} & & Subject Imprecision & & & 0.6454 & 0.6423 &  0.2647 / 0.1607 / 0.75  &  0.1695 / 0.0962 / 0.7143  \\
\cline{2-9} & \multirow{7}{*}{ Search QA } & Incorrect Answer/Unrelated Matching Results & \multirow{7}{*}{\begin{tabular}{l}
\begin{tabular}{l}
0.5705 \\
0.5398 \\
0.5226
\end{tabular}
\end{tabular}} & \multirow{7}{*}{\begin{tabular}{l}
\begin{tabular}{l}
0.4772 \\
0.4553 \\
0.4425
\end{tabular}
\end{tabular}} & 0.748 & 0.7934 &  0.2381 / 0.1852 / 0.3333  &  0.4186 / 0.3462 / 0.5294  \\
\cline{3-3} \cline{6-9} & & External Links or Diversions & & & 0.8425 & 0.8678 &  0.3333 / 0.2273 / 0.625  &  0.2 / 0.1333 / 0.4  \\
\cline{3-3} \cline{6-9} & & Refusal to Answer & & & 0.9213 & 0.9504 & 0.5/0.3333/1.0 & 0.5/0.4286/0.6  \\
\cline{3-3} \cline{6-9} & &  Incomplete Answers  & & & 0.6693 & 0.7273 &  0.125 / 0.0882 / 0.2143  &  0.2979 / 0.2188 / 0.4667  \\
\cline{3-3} \cline{6-9} & & Stiffness & & & 0.7559 & 0.6694 &  0.0606 / 0.0323 / 0.5  & 0.1667/0.093/0.8  \\
\cline{3-3} \cline{6-9} & & Repetitive Expression & & & 0.4803 & 0.4793 &  0.0833 / 0.0435 / 1.0  &  0.0597 / 0.0308 / 1.0  \\
\cline{3-3} \cline{6-9} & & Subject Imprecision & & & 0.7323 & 0.8017 &  0.0556 / 0.0312 / 0.25  &  0.25 / 0.2 / 0.3333  \\
\cline{2-9} & \multirow{8}{*}{ Title Generation } & Not Meeting the Requirements & \multirow{8}{*}{\begin{tabular}{l}
\begin{tabular}{l}
0.2738 \\
0.2746 \\
0.2607
\end{tabular}
\end{tabular}} & \multirow{8}{*}{\begin{tabular}{l}
\begin{tabular}{l}
0.2671 \\
0.2634 \\
0.2546
\end{tabular}
\end{tabular}} & 0.5915 & 0.6537 &  0.6098 / 0.9036 / 0.4601  &  0.685 / 0.9667 / 0.5305  \\
\cline{3-3} \cline{6-9} & & Incorrect Answer/Unrelated Matching Results & & & 0.7021 & 0.6883 &  0.4355 / 0.3699 / 0.5294  &  0.4706 / 0.4571 / 0.4848   \\
\cline{3-3} \cline{6-9} & & External Links or Diversion & & & 0.8298 & 0.8745 &  0.1304 / 0.0698 / 1.0  &  0.1714 / 0.0938 / 1.0 \\
\cline{3-3} \cline{6-9} & &  Untranslated Text & & & 0.7745 & 0.7792 &  0.1311 / 0.0727 / 0.6667  &  0.1639 / 0.0926 / 0.7143   \\
\cline{3-3} \cline{6-9} & & Confusing Answers & & & 0.8511 & 0.8442 &  0.2222 / 0.1429 / 0.5  &  0.1818 / 0.1081 / 0.5714    \\
\cline{3-3} \cline{6-9} & & Stiffness & & & 0.7021 & 0.71 &  0.1463 / 0.0811 / 0.75  &  0.2947 / 0.2 / 0.56  \\
\cline{3-3} \cline{6-9} & & Repetitive Expression & & & 0.4638 & 0.4545 &  0.1 / 0.0534 / 0.7778  &  0.0735 / 0.0382 / 1.0   \\
\cline{3-3} \cline{6-9} & & Subject Imprecision & & & 0.7191 & 0.7403 &  0.2979 / 0.2295 / 0.4242  &  0.3023 / 0.2241 / 0.4643  \\
\hline
\end{tabular}
}
\caption{The overall results of Qwen-72B-Chat + SFT}
\label{tab:tab104}
\end{table*}

\begin{table*}
\centering
\scalebox{0.5}{
\begin{tabular}{|c|c|c|c|c|c|c|c|c|}
\hline \multirow{2}{*}{ Method } & \multirow{2}{*}{ task } & \multirow{2}{*}{ label } & \multicolumn{2}{|c|}{ Spearman/Pearson/Kendall } & \multicolumn{2}{|c|}{ Acc } & \multicolumn{2}{|c|}{ F1/Precision/Recall } \\
\cline{4-9} & & & eval & test & eval & test & eval & test \\
\hline \multirow{31}{*}{ \shortstack{Qwen-72B-Chat \\ + ICL}} & \multirow{8}{*}{\begin{tabular}{l} 
Sentiment Analysis
\end{tabular}} & Not Meeting the Requirements & \multirow{8}{*}{\begin{tabular}{l}
\begin{tabular}{l}
0.7676 \\
0.7443 \\
0.7122
\end{tabular}
\end{tabular}} & \multirow{8}{*}{\begin{tabular}{l}
\begin{tabular}{l}
0.7578 \\
0.7303 \\
0.7128
\end{tabular}
\end{tabular}} & 0.9298 & 0.8649 & 0.75/0.75/0.75 &  0.4828 / 0.4118 / 0.5833  \\
\cline{3-3} \cline{6-9} & & Incorrect Answer/Unrelated Matching Results & & & 0.9035 & 0.8649 &  0.8791 / 0.9524 / 0.8163  &  0.8485 / 0.875 / 0.8235  \\
\cline{3-3} \cline{6-9} & &  Refusal to Answer   & & & 0.9561 & 0.9369 & 0.6154/0.5/0.8 &  0.3636 / 0.2222 / 1.0  \\
\cline{3-3} \cline{6-9} & & Untranslated Text & & & 0.9649 & 0.964 &  0.7778 / 0.6364 / 1.0  & 0.75/0.6667/0.8571  \\
\cline{3-3} \cline{6-9} & &  Confusing Answers  & & & 0.8596 & 0.8378 &  0.2 / 0.1176 / 0.6667  &  0.1 / 0.0588 / 0.3333  \\
\cline{3-3} \cline{6-9} & & Stiffness & & & 0.7895 & 0.7477 &  0.2941 / 0.2 / 0.5556  &  0.3 / 0.1875 / 0.75  \\
\cline{3-3} \cline{6-9} & &  Repetitive Expression  & & & 0.8246 & 0.8378 &  0.2308 / 0.1765 / 0.3333  &  0.3077 / 0.2222 / 0.5  \\
\cline{3-3} \cline{6-9} & &  Subject Imprecision  & & & 0.5439 & 0.6126 &  0.3158 / 0.24 / 0.4615  &  0.1569 / 0.0909 / 0.5714  \\
\cline{2-9} & \multirow{8}{*}{ Knowledge QA } &  Incorrect Answer/Unrelated Matching Results  & \multirow{8}{*}{\begin{tabular}{l}
\begin{tabular}{l}
0.2969 \\
0.301 \\
0.2747
\end{tabular}
\end{tabular}} & \multirow{8}{*}{\begin{tabular}{l}
\begin{tabular}{l}
0.1766 \\
0.1988 \\
0.1653
\end{tabular}
\end{tabular}} & 0.6099 & 0.4818 &  0.3038 / 0.1846 / 0.8571  &  0.1647 / 0.0946 / 0.6364  \\
\cline{3-3} \cline{6-9} & &  Refusal to Answer   & & & 0.9433 & 0.9416 &  0.4286 / 0.2727 / 1.0  &  0.4286 / 0.2727 / 1.0   \\
\cline{3-3} \cline{6-9} & &  Untranslated Text  & & & 0.9645 & 0.9562 &  0.6154 / 0.4444 / 1.0  & 0.5/0.375/0.75  \\
\cline{3-3} \cline{6-9} & &  Confusing Answers  & & & 0.9007 & 0.9197 &  0.2222 / 0.1333 / 0.6667  &  0.2667 / 0.2 / 0.4 \\
\cline{3-3} \cline{6-9} & & Incomplete Answers & & & 0.7376 & 0.7372 &  0.6542 / 0.6604 / 0.6481  &  0.7143 / 0.7377 / 0.6923  \\
\cline{3-3} \cline{6-9} & &  Stiffness  & & & 0.6525 & 0.6715 &  0.2462 / 0.1404 / 1.0  &  0.2105 / 0.12 / 0.8571 \\
\cline{3-3} \cline{6-9} & & Repetitive Expression & & & 0.8298 & 0.8759 &  0.2941 / 0.1852 / 0.7143  &  0.2609 / 0.1765 / 0.5   \\
\cline{3-3} \cline{6-9} & & Subject Imprecision & & & 0.8511 & 0.8102 &  0.4615 / 0.3333 / 0.75  &  0.2353 / 0.1481 / 0.5714  \\
\cline{2-9} & \multirow{7}{*}{ Search QA } & Incorrect Answer/Unrelated Matching Results & \multirow{7}{*}{\begin{tabular}{l}
\begin{tabular}{l}
0.519 \\
0.5028 \\
0.4849
\end{tabular}
\end{tabular}} & \multirow{7}{*}{\begin{tabular}{l}
\begin{tabular}{l}
0.4513 \\
0.437 \\
0.4238
\end{tabular}
\end{tabular}} & 0.811 & 0.8512 &  0.3333 / 0.2857 / 0.4  &  0.4706 / 0.4706 / 0.4706  \\
\cline{3-3} \cline{6-9} & & External Links or Diversions & & & 0.8583 & 0.8843 &  0.25 / 0.1875 / 0.375  &  0.2222 / 0.1538 / 0.4  \\
\cline{3-3} \cline{6-9} & & Refusal to Answer & & & 0.8504 & 0.8264 &  0.3448 / 0.2083 / 1.0  &  0.3226 / 0.1923 / 1.0  \\
\cline{3-3} \cline{6-9} & &  Incomplete Answers  & & & 0.5354 & 0.6033 &  0.2338 / 0.1429 / 0.6429  &  0.3684 / 0.2295 / 0.9333  \\
\cline{3-3} \cline{6-9} & & Stiffness & & & 0.6929 & 0.7851 &  0.0488 / 0.0256 / 0.5  &  0.2353 / 0.1379 / 0.8  \\
\cline{3-3} \cline{6-9} & & Repetitive Expression & & & 0.6614 & 0.7107 &  0.1224 / 0.0652 / 1.0  &  0.1026 / 0.0541 / 1.0   \\
\cline{3-3} \cline{6-9} & & Subject Imprecision & & & 0.685 & 0.7273 &  0.0476 / 0.0263 / 0.25  &  0.2979 / 0.2 / 0.5833  \\
\cline{2-9} & \multirow{8}{*}{ Title Generation } & Not Meeting the Requirements & \multirow{8}{*}{\begin{tabular}{l}
\begin{tabular}{l}
0.1585 \\
0.156 \\
0.1521
\end{tabular}
\end{tabular}} & \multirow{8}{*}{\begin{tabular}{l}
\begin{tabular}{l}
0.1392 \\
0.1021 \\
0.1337
\end{tabular}
\end{tabular}} & 0.7617 & 0.7576 &  0.8303 / 0.8204 / 0.8405  &  0.8313 / 0.8214 / 0.8415  \\
\cline{3-3} \cline{6-9} & & Incorrect Answer/Unrelated Matching Results & & & 0.6596 & 0.7749 &  0.4737 / 0.3564 / 0.7059  &  0.6667 / 0.5778 / 0.7879   \\
\cline{3-3} \cline{6-9} & & External Links or Diversion & & & 0.9915 & 0.987 &  0.75 / 0.6 / 1.0  &  0.5714 / 0.5 / 0.6667 \\
\cline{3-3} \cline{6-9} & &  Untranslated Text & & & 0.634 & 0.6147 &  0.1042 / 0.0556 / 0.8333  &  0.1359 / 0.0729 / 1.0   \\
\cline{3-3} \cline{6-9} & & Confusing Answers & & & 0.6723 & 0.6753 &  0.1348 / 0.0759 / 0.6  &  0.1176 / 0.0641 / 0.7143    \\
\cline{3-3} \cline{6-9} & & Stiffness & & & 0.8298 & 0.7879 &  0.1304 / 0.0789 / 0.375  &  0.2222 / 0.1842 / 0.28  \\
\cline{3-3} \cline{6-9} & & Repetitive Expression & & & 0.6043 & 0.5714 &  0.0971 / 0.0532 / 0.5556  &  0.0571 / 0.03 / 0.6  \\
\cline{3-3} \cline{6-9} & & Subject Imprecision & & & 0.6979 & 0.684 &  0.297 / 0.2206 / 0.4545  &  0.2913 / 0.2 / 0.5357  \\
\hline
\end{tabular}
}
\caption{The overall results of Qwen-72B-Chat + ICL}
\label{tab:tab105}
\end{table*}

\begin{table*}
\centering
\scalebox{0.5}{
\begin{tabular}{|c|c|c|c|c|c|c|c|c|}
\hline \multirow{2}{*}{ Method } & \multirow{2}{*}{ task } & \multirow{2}{*}{ label } & \multicolumn{2}{|c|}{ Spearman/Pearson/Kendall } & \multicolumn{2}{|c|}{ Acc } & \multicolumn{2}{|c|}{ F1/Precision/Recall } \\
\cline{4-9} & & & eval & test & eval & test & eval & test \\
\hline \multirow{31}{*}{ Non-division } & \multirow{8}{*}{\begin{tabular}{l} 
Sentiment Analysis
\end{tabular}} & Not Meeting the Requirements & \multirow{8}{*}{\begin{tabular}{l}
\begin{tabular}{l}
0.3735 \\
0.3856 \\
0.3631
\end{tabular}
\end{tabular}} & \multirow{8}{*}{\begin{tabular}{l}
\begin{tabular}{l}
0.2905 \\
0.317 \\
0.2838
\end{tabular}
\end{tabular}} & 0.8333 & 0.8559 &  0.4571 / 0.4211 / 0.5  &  0.3846 / 0.3571 / 0.4167  \\
\cline{3-3} \cline{6-9} & & Incorrect Answer/Unrelated Matching Results & & & 0.6404 & 0.5676 &  0.3279 / 0.8333 / 0.2041  &  0.2 / 0.6667 / 0.1176  \\
\cline{3-3} \cline{6-9} & &  Refusal to Answer   & & & 0.9561 & 0.973 &  0.5455 / 0.5 / 0.6  &  0.5714 / 0.4 / 1.0  \\
\cline{3-3} \cline{6-9} & & Untranslated Text & & & 0.9912 & 0.973 &  0.9333 / 0.875 / 1.0  &  0.8 / 0.75 / 0.8571  \\
\cline{3-3} \cline{6-9} & &  Confusing Answers  & & & 0.9649 & 0.964 & 0.6/0.4286/1.0 & 0.6/0.4286/1.0  \\
\cline{3-3} \cline{6-9} & & Stiffness & & & 0.9298 & 0.9459 &  0.4286 / 0.6 / 0.3333  &  0.6667 / 0.6 / 0.75  \\
\cline{3-3} \cline{6-9} & &  Repetitive Expression  & & & 0.9386 & 0.9279 &  0.4615 / 0.75 / 0.3333  &  0.3333 / 0.5 / 0.25  \\
\cline{3-3} \cline{6-9} & &  Subject Imprecision  & & & 0.807 & 0.9009 &  0.3889 / 0.7 / 0.2692  &  0.1538 / 0.1667 / 0.1429  \\
\cline{2-9} & \multirow{8}{*}{ Knowledge QA } &  Incorrect Answer/Unrelated Matching Results  & \multirow{8}{*}{\begin{tabular}{l}
\begin{tabular}{l}
0.5398 \\
0.5877 \\
0.5173
\end{tabular}
\end{tabular}} & \multirow{8}{*}{\begin{tabular}{l}
\begin{tabular}{l}
0.4251 \\
0.4361 \\
0.406
\end{tabular}
\end{tabular}} & 0.9149 & 0.8321 &  0.5714 / 0.5714 / 0.5714  &  0.1481 / 0.125 / 0.1818  \\
\cline{3-3} \cline{6-9} & &  Refusal to Answer   & & & 0.9504 & 0.9635 &  0.3636 / 0.25 / 0.6667  &  0.5455 / 0.375 / 1.0   \\
\cline{3-3} \cline{6-9} & &  Untranslated Text  & & & 0.9787 & 0.9781 &  0.7273 / 0.5714 / 1.0  &  0.7273 / 0.5714 / 1.0  \\
\cline{3-3} \cline{6-9} & &  Confusing Answers  & & & 0.9574 & 0.9708 &  0.4 / 0.2857 / 0.6667  &  0.7143 / 0.5556 / 1.0 \\
\cline{3-3} \cline{6-9} & & Incomplete Answers & & & 0.695 & 0.6058 & 0.411/0.7895/0.2778 &  0.3721 / 0.7619 / 0.2462  \\
\cline{3-3} \cline{6-9} & &  Stiffness  & & & 0.9362 & 0.9197 &  0.4706 / 0.4444 / 0.5  &  0.3529 / 0.3 / 0.4286 \\
\cline{3-3} \cline{6-9} & & Repetitive Expression & & & 0.9291 & 0.927 &  0.375 / 0.3333 / 0.4286  &  0.1667 / 0.1667 / 0.1667   \\
\cline{3-3} \cline{6-9} & & Subject Imprecision & & & 0.8936 & 0.8905 &  0.4 / 0.3846 / 0.4167  &  0.1176 / 0.1 / 0.1429  \\
\cline{2-9} & \multirow{7}{*}{ Search QA } & Incorrect Answer/Unrelated Matching Results & \multirow{7}{*}{\begin{tabular}{l}
\begin{tabular}{l}
0.689 \\
0.6867 \\
0.6587
\end{tabular}
\end{tabular}} & \multirow{7}{*}{\begin{tabular}{l}
\begin{tabular}{l}
0.4691 \\
0.4693 \\
0.4529
\end{tabular}
\end{tabular}} & 0.7795 & 0.6777 &  0.3636 / 0.2759 / 0.5333  &  0.2353 / 0.1765 / 0.3529  \\
\cline{3-3} \cline{6-9} & & External Links or Diversions & & & 0.8425 & 0.7686 &  0.3333 / 0.2273 / 0.625  &  0.1765 / 0.1034 / 0.6  \\
\cline{3-3} \cline{6-9} & & Refusal to Answer & & & 0.8031 & 0.7438 &  0.2424 / 0.1429 / 0.8  &  0.1622 / 0.0938 / 0.6  \\
\cline{3-3} \cline{6-9} & &  Incomplete Answers  & & & 0.7165 & 0.7107 &  0.2174 / 0.1562 / 0.3571  &  0.3137 / 0.2222 / 0.5333  \\
\cline{3-3} \cline{6-9} & & Stiffness & & & 0.8031 & 0.7107 &  0.1379 / 0.0741 / 1.0  &  0.0541 / 0.0312 / 0.2  \\
\cline{3-3} \cline{6-9} & & Repetitive Expression & & & 0.8189 & 0.7355 &  0.1481 / 0.0833 / 0.6667  & 0/0/0   \\
\cline{3-3} \cline{6-9} & & Subject Imprecision & & & 0.748 & 0.7025 &  0.1111 / 0.0625 / 0.5  &  0.25 / 0.1667 / 0.5    \\
\cline{2-9} & \multirow{8}{*}{ Title Generation } & Not Meeting the Requirements & \multirow{8}{*}{\begin{tabular}{l}
\begin{tabular}{l}
0.3763 \\
0.3701 \\
0.3556
\end{tabular}
\end{tabular}} & \multirow{8}{*}{\begin{tabular}{l}
\begin{tabular}{l}
0.4296 \\
0.4443 \\
0.4051
\end{tabular}
\end{tabular}} & 0.6468 & 0.6364 &  0.6844 / 0.9 / 0.5521  &  0.6866 / 0.8846 / 0.561  \\
\cline{3-3} \cline{6-9} & & Incorrect Answer/Unrelated Matching Results & & & 0.7532 & 0.7359 &  0.383 / 0.4186 / 0.3529  &  0.4404 / 0.5581 / 0.3636   \\
\cline{3-3} \cline{6-9} & & External Links or Diversion & & & 0.8979 & 0.8961 &  0.2 / 0.1111 / 1.0  &  0.2 / 0.1111 / 1.0 \\
\cline{3-3} \cline{6-9} & &  Untranslated Text & & & 0.8979 & 0.8874 &  0.25 / 0.1538 / 0.6667  &  0.2353 / 0.1481 / 0.5714   \\
\cline{3-3} \cline{6-9} & & Confusing Answers & & & 0.8 & 0.7403 &  0.1754 / 0.1064 / 0.5  &  0.0909 / 0.0508 / 0.4286    \\
\cline{3-3} \cline{6-9} & & Stiffness & & & 0.8043 & 0.7403 &  0.1786 / 0.1042 / 0.625  &  0.25 / 0.1818 / 0.4  \\
\cline{3-3} \cline{6-9} & & Repetitive Expression & & & 0.6851 & 0.6147 &  0.1395 / 0.0779 / 0.6667  &  0.0632 / 0.0333 / 0.6  \\
\cline{3-3} \cline{6-9} & & Subject Imprecision & & & 0.783 & 0.8009 &  0.2154 / 0.2188 / 0.2121  &  0.2581 / 0.2353 / 0.2857  \\
\hline
\end{tabular}
}
\caption{The overall results of Non-division}
\label{tab:tab106}
\end{table*}

\begin{table*}
\centering
\scalebox{0.5}{
\begin{tabular}{|c|c|c|c|c|c|c|c|c|}
\hline \multirow{2}{*}{ Method } & \multirow{2}{*}{ task } & \multirow{2}{*}{ label } & \multicolumn{2}{|c|}{ Spearman/Pearson/Kendall } & \multicolumn{2}{|c|}{ Acc } & \multicolumn{2}{|c|}{ F1/Precision/Recall } \\
\cline{4-9} & & & eval & test & eval & test & eval & test \\
\hline \multirow{31}{*}{ \shortstack{ Non-repetition } } & \multirow{8}{*}{\begin{tabular}{l} 
Sentiment Analysis
\end{tabular}} & Not Meeting the Requirements & \multirow{8}{*}{\begin{tabular}{l}
\begin{tabular}{l}
0.9553 \\
0.9536 \\
0.9495
\end{tabular}
\end{tabular}} & \multirow{8}{*}{\begin{tabular}{l}
\begin{tabular}{l}
0.9658 \\
0.953 \\
0.9515
\end{tabular}
\end{tabular}} & 0.9386 & 0.8829 &  0.7879 / 0.7647 / 0.8125  &  0.5517 / 0.4706 / 0.6667  \\
\cline{3-3} \cline{6-9} & & Incorrect Answer/Unrelated Matching Results & & & 0.9825 & 0.9189 &  0.9792 / 1.0 / 0.9592  &  0.9091 / 0.9375 / 0.8824  \\
\cline{3-3} \cline{6-9} & &  Refusal to Answer   & & & 0.9737 & 0.982 &  0.6667 / 0.75 / 0.6  &  0.6667 / 0.5 / 1.0  \\
\cline{3-3} \cline{6-9} & & Untranslated Text & & & 1 & 0.982 & 1.0/1.0/1.0 &  0.875 / 0.7778 / 1.0  \\
\cline{3-3} \cline{6-9} & &  Confusing Answers  & & & 1 & 0.991 & 1.0/1.0/1.0 &  0.8571 / 0.75 / 1.0  \\
\cline{3-3} \cline{6-9} & & Stiffness & & & 0.9825 & 0.973 &  0.9 / 0.8182 / 1.0  &  0.8421 / 0.7273 / 1.0  \\
\cline{3-3} \cline{6-9} & &  Repetitive Expression  & & & 0.9561 & 0.9459 &  0.6154 / 1.0 / 0.4444  &  0.4 / 1.0 / 0.25  \\
\cline{3-3} \cline{6-9} & &  Subject Imprecision  & & & 0.8772 & 0.8018 &  0.7586 / 0.6875 / 0.8462  &  0.3529 / 0.2222 / 0.8571  \\
\cline{2-9} & \multirow{8}{*}{ Knowledge QA } &  Incorrect Answer/Unrelated Matching Results  & \multirow{8}{*}{\begin{tabular}{l}
\begin{tabular}{l}
0.4911 \\
0.506 \\
0.4721
\end{tabular}
\end{tabular}} & \multirow{8}{*}{\begin{tabular}{l}
\begin{tabular}{l}
0.4823 \\
0.4784 \\
0.4603
\end{tabular}
\end{tabular}} & 0.8511 & 0.7883 &  0.5333 / 0.3871 / 0.8571  &  0.2927 / 0.2 / 0.5455  \\
\cline{3-3} \cline{6-9} & &  Refusal to Answer   & & & 0.9929 & 0.9854 &  0.8571 / 0.75 / 1.0  &  0.75 / 0.6 / 1.0   \\
\cline{3-3} \cline{6-9} & &  Untranslated Text  & & & 0.9929 & 1 &  0.8889 / 0.8 / 1.0  &  1.0 / 1.0 / 1.0  \\
\cline{3-3} \cline{6-9} & &  Confusing Answers  & & & 0.9858 & 0.9927 &  0.75 / 0.6 / 1.0  & 0.8889/1.0/0.8 \\
\cline{3-3} \cline{6-9} & & Incomplete Answers & & & 0.8014 & 0.7518 &  0.7021 / 0.825 / 0.6111  &  0.6667 / 0.9189 / 0.5231  \\
\cline{3-3} \cline{6-9} & &  Stiffness  & & & 0.9716 & 1 &  0.7143 / 0.8333 / 0.625  & 1.0/1.0/1.0 \\
\cline{3-3} \cline{6-9} & & Repetitive Expression & & & 0.9929 & 0.9635 &  0.9231 / 1.0 / 0.8571  &  0.6154 / 0.5714 / 0.6667   \\
\cline{3-3} \cline{6-9} & & Subject Imprecision & & & 0.9504 & 0.9416 &  0.6667 / 0.7778 / 0.5833  &  0.2 / 0.3333 / 0.1429  \\
\cline{2-9} & \multirow{7}{*}{ Search QA } & Incorrect Answer/Unrelated Matching Results & \multirow{7}{*}{\begin{tabular}{l}
\begin{tabular}{l}
0.8208 \\
0.8163 \\
0.7961
\end{tabular}
\end{tabular}} & \multirow{7}{*}{\begin{tabular}{l}
\begin{tabular}{l}
0.838 \\
0.838 \\
0.8304
\end{tabular}
\end{tabular}} & 0.9606 & 0.9669 &  0.8148 / 0.9167 / 0.7333  &  0.8667 / 1.0 / 0.7647  \\
\cline{3-3} \cline{6-9} & & External Links or Diversions & & & 1 & 0.9752 &  1.0 / 1.0 / 1.0  &  0.7273 / 0.6667 / 0.8  \\
\cline{3-3} \cline{6-9} & & Refusal to Answer & & & 0.9843 & 0.9917 &  0.8333 / 0.7143 / 1.0  & 0.8889/1.0/0.8  \\
\cline{3-3} \cline{6-9} & &  Incomplete Answers  & & & 0.8898 & 0.9091 &  0.4167 / 0.5 / 0.3571  &  0.4762 / 0.8333 / 0.3333  \\
\cline{3-3} \cline{6-9} & & Stiffness & & & 0.9764 & 0.9752 &  0.5714 / 0.4 / 1.0  &  0.5714 / 1.0 / 0.4  \\
\cline{3-3} \cline{6-9} & & Repetitive Expression & & & 0.9843 & 0.9835 &  0.6667 / 0.6667 / 0.6667  &  0.6667 / 0.5 / 1.0   \\
\cline{3-3} \cline{6-9} & & Subject Imprecision & & & 0.937 & 0.9421 &  0.4286 / 0.3 / 0.75  &  0.6316 / 0.8571 / 0.5    \\
\cline{2-9} & \multirow{8}{*}{ Title Generation } & Not Meeting the Requirements & \multirow{8}{*}{\begin{tabular}{l}
\begin{tabular}{l}
0.6618 \\
0.6716 \\
0.6499
\end{tabular}
\end{tabular}} & \multirow{8}{*}{\begin{tabular}{l}
\begin{tabular}{l}
0.6815 \\
0.6663 \\
0.6673
\end{tabular}
\end{tabular}} & 0.8426 & 0.8615 &  0.881 / 0.9257 / 0.8405  &  0.8974 / 0.9459 / 0.853  \\
\cline{3-3} \cline{6-9} & & Incorrect Answer/Unrelated Matching Results & & & 0.9362 & 0.8961 &  0.8515 / 0.86 / 0.8431  &  0.7966 / 0.9038 / 0.7121   \\
\cline{3-3} \cline{6-9} & & External Links or Diversion & & & 1 & 1 & 1.0/1.0/1.0 & 1.0/1.0/1.0 \\
\cline{3-3} \cline{6-9} & &  Untranslated Text & & & 0.9489 & 0.987 &  0.5 / 0.3333 / 1.0  &  0.8 / 0.75 / 0.8571   \\
\cline{3-3} \cline{6-9} & & Confusing Answers & & & 0.9404 & 0.9654 &  0.5625 / 0.4091 / 0.9  &  0.6364 / 0.4667 / 1.0    \\
\cline{3-3} \cline{6-9} & & Stiffness & & & 0.9574 & 0.9784 &  0.6154 / 0.4444 / 1.0  &  0.898 / 0.9167 / 0.88  \\
\cline{3-3} \cline{6-9} & & Repetitive Expression & & & 0.966 & 0.9784 &  0.6364 / 0.5385 / 0.7778  &  0.6667 / 0.5 / 1.0  \\
\cline{3-3} \cline{6-9} & & Subject Imprecision & & & 0.9064 & 0.9221 &  0.7179 / 0.6222 / 0.8485  &  0.7 / 0.6562 / 0.75  \\
\hline
\end{tabular}
}
\caption{The overall results of Standard Prompt Paradigm(without repetition)}
\label{tab:tab107}
\end{table*}

\end{document}